\documentclass[journal]{IEEEtranTIE}

\ifCLASSINFOpdf
\else
\fi
\usepackage[linesnumbered, ruled]{algorithm2e}
\usepackage{setspace}

\usepackage{amsmath}
\usepackage{graphicx}
\usepackage{times}
\usepackage{epsfig}
\usepackage{amssymb}
\usepackage{mathrsfs}
\usepackage[table,xcdraw]{xcolor}
\usepackage{color}
\usepackage[hidelinks]{hyperref}
\usepackage{url}
\usepackage{siunitx}
\usepackage{booktabs}
\usepackage{tabularx}
\usepackage{float}
\usepackage{threeparttable}
\usepackage{diagbox}%斜线表头
\usepackage{multirow}
\usepackage{amsfonts}

\newcommand{\gou}{$\checkmark$}
\newcommand{\wu}{$-	$}

\newcommand{\revise}{}
\def\degree{${}^{\circ}$}
\usepackage{tikz}
\def \

\ifCLASSINFOpdf

\else

\fi

\begin{document}
	\title{Multiview Point Cloud Registration Based on Minimum Potential Energy for Free-Form Blade Measurement}
	%Minimum Potential Energy of Point Cloud for Robust Global Registration
	\author{Zijie Wu,~Yaonan Wang,~Yang Mo,~Qing Zhu,~He Xie,~Haotian Wu,~Mingtao Feng and Ajmal Mian
			%\author{Zijie Wu,~Yaonan Wang,~Qing Zhu,~Jingmou Nie,~Mingtao Feng and Ajmal Mian
		%\author{Zijie Wu,~Yaonan Wang,~Qing Zhu,~Jianxu Mao,~Haotian Wu and Mingtao Feng	

		\thanks{Z. Wu, Y. Wang, Y. Mo, Q. Zhu, H. Xie, and H. Wu are with the College of Electrical and Information Engineering, Hunan University, Changsha 410082, China (email:wuzijieeee, yaonan, moyanghnu, zhuqing, xiehe, wuhaotian@hnu.edu.cn).\emph{(Corresponding author:
        Yang Mo.)}}% <-this % stops a space
		
		\thanks{M. Feng is with the School of Computer Science and Technology, Xidian University, Xi’an 710071, China (email: mintfeng@hnu.edu.cn).}% <-this % stops a space
		
		\thanks{A. Mian is with the Department of Computer Science and Software Engineering, The University of Western Australia, Perth, Crawley, WA 6009, Australia (e-mail: ajmal.mian@uwa.edu.au).}% <-this % stops a space
		
		\thanks{This work was supported in part by the National Natural Science Foundation of China under grants 61733004, 62076091, and 62003253, in part by the China National Postdoctoral Program for Innovative Talents under grant No. BX2021098 and 2021M701155, in part by the Major Project of Science and Technology Innovation 2030 under grant No.2021ZD0113100, in part by Natural Science Foundation of Changsha under grant kq2202136 and kq2014056, and in part by the Australian Research Council under grant DP190102443.}
		
		%\thanks{Manuscript received ***; revised ***.}
		}
	
	\maketitle
	
	\begin{abstract}
Point cloud registration is an essential step for free-form blade reconstruction in industrial measurement.
Nonetheless, measuring defects of the 3D acquisition system unavoidably result in noisy and incomplete point cloud data, which renders efficient and accurate registration challenging. In this paper, we propose a novel global registration method that is based on the minimum potential energy (MPE) method to address these problems. The basic strategy is that the objective function is defined as the minimum potential energy optimization function of the physical registration system. The function distributes more weight to the majority of inlier points and less weight to the noise and outliers, which essentially reduces the influence of perturbations in the mathematical formulation.
We decompose the solution into a globally optimal approximation procedure and a fine registration process with the trimmed iterative closest point algorithm to boost convergence. The approximation procedure consists of two main steps. First, according to the construction of the force traction operator, we can simply compute the position of the potential energy minimum. Second, to find the MPE point, we propose a new theory that employs two flags to observe the status of the registration procedure. 
We demonstrate the performance of the proposed algorithm on four types of blades. The proposed method outperforms the other global methods in terms of both accuracy and noise resistance.
	\end{abstract}
	
	\begin{IEEEkeywords}
		Free-form blade measurement, minimum energy registration, measuring defects, global robust method, turbine blade
	\end{IEEEkeywords}
	
	%%方法微改，参考文献修改。

	% \fi
	%
	% For peer review papers, this IEEEtran command inserts a page break and
	\IEEEpeerreviewmaketitle

	\section{Introduction}
	\IEEEPARstart{T}{he} free-form blade is a fundamental structure of industrial components and is often used in precise aeroengines and steam turbines. %??????????????\revise{}
	Traditional blade measurement~\cite{CMM1} aims to produce a model using a coordinate measuring machine (CMM). The high precision of CMMs makes them suitable for application in final inspection. However, limited by the performance of the measurement machine, it is usually time-consuming with low productivity. Optical measurement methods of other types are widely used for object measurement, such as stereo vision using surface structured light~\cite{CMM2}. This method realizes higher efficiency and obtains precise point cloud data for the quality evaluation of blades. 
	\revise{Based on stereo scans, a more common approach for correspondence establishment between different views is proposed. It depends on artificial markers, which are manually placed on the measured surface before scans. The common marker points that are visible between the aligned scans are extracted either manually~\cite{marker2} or automatically~\cite{marker1} and then the transformation parameters for registration are determined. Marker point-based methods show precise performance if the manually placed markers remain stable during the measurement. However, several characteristics of marker point-method make them difficult to apply for automatic measurement in blade manufacturing: 1) The manual intervention of marker placement is usually time-consuming because the markers need to be distributed stably to avoid ill-defined geometric constellations. The time cost is especially tremendous when dealing with mass-produced workpieces. 2) In automatic manufacturing, the marker points are vulnerable during processes such as milling and grinding. In most situations, marker point-based methods are applied in an offline measurement when the blade is not simultaneously required to be processed.}%%%
		
	\begin{table*}[htbp]
		\centering
		\addtolength{\tabcolsep}{0pt}
		\caption{\label{tab:12}\revise{Characteristics of Three Types of Measurement Methods}}
		\begin{tabular}{c| c c c}
			\toprule
			Characteristic				& CMM & Markers & PCR \\
			\midrule
			Manual Intervention				& $-$ & $\checkmark$ & $-	$\\
			Speed							& Extremely Slow & Slow & Fast  \\
			Application						& Final Inspection & Offline Measurement & Online Measurement  \\ 
			%Repeatability					& 5.1m & 3.9s & 1.5s \\ 
			\bottomrule
		\end{tabular}
	\end{table*}	
	
	\revise{To circumvent the manual intervention requirement and the vulnerable marker points, we apply direct 3D point cloud registration (PCR) for automatic and accurate blade measurement. It can be used in the applications during processing such as automatic margin calculation in grinding and milling.
	The overall characteristics are presented in Table~\ref{tab:12}. We consider three characteristics: the manual intervention, measurement speed, and application scenarios. In terms of all these characteristics, manual intervention imposes higher requirements on operators and results in lower consistency. We consider the time-cost for measurement and the target application scenarios. The PCR technique is considered most suitable for repeated measurements during blade processing in industrial manufacturing. Precise blade reconstruction to meet the requirements of the measurement system using PCR is the main problem to be solved in this paper.
	}%%系统是咋干的。%%加一点要求的叶片测量精度
	Nonetheless, limited by the single-scan view of the camera, the surface structured light scheme hardly establishes the correspondence between the point clouds of multiple view scans and necessitates registration with accurate methods~\cite{lei2017fast} for model reconstruction.
	
%	\begin{table}[htbp]
%		\centering
%		\addtolength{\tabcolsep}{-5pt}
%		\caption{\label{tab:12}The Time Cost of Minimum PE Registration For The Naive MPE Theory and Two Down-Sampling Rates of The MPL Algorithm}
%		\begin{tabular}{c c c c c}
%			\toprule
%			Methods 				& Human Intervention & Speed & Accuracy & Repeatability\\
%			\midrule
%			CMM 					& MPE (full) & MPL (200) & MPL (100)&\\
%			Markers			& 16.4s & 3.3s & 0.9s & \\
%			PCR						& 5.1m & 3.9s & 1.5s & \\ 
%			\bottomrule
%		\end{tabular}
%	\end{table}

	The 3D point cloud registration~\cite{tim1} technique is an important procedure for 3D blade reconstruction~\cite{peng2020viewpoints}~\cite{ef2}, which reconstructs measured blade workpieces~\cite{tim3} for comparison with a standard model for quality evaluation~\cite{tim2}. Three-dimensional point cloud registration is the task of establishing correspondences between point clouds that have been scanned from different views, each residing in a different coordinate of a camera system, and subsequently minimizing the distances between the corresponding point pairs to align the point clouds in the same coordinate system. Therefore, the precise reconstruction of a free-form workpiece in industrial applications necessitates an accurate registration algorithm. In recent years, the ubiquity of 3D acquisition devices in the manufacturing area has led to a growing interest in the high-precision reconstruction of workpieces and the need for more effective, robust and efficient algorithms. However, 3D acquisition devices, especially those that operate in real time, provide noisy point cloud data, which makes it challenging to achieve efficient and accurate registration. To solve this problem, we propose a novel global registration method that is based on the minimum potential energy method for addressing the problem of noise, including measurement noise and outliers. It assigns different weights to the inlier part and noise part using a criterion function and realizes fine registration for blade measurement.

	%	\begin{figure}[t!] 
	%	\center{\includegraphics[width=0.5\textwidth]{fig4.png}}
	%	\vspace{-5mm}
	%	\caption{\label{fig:fig4} Reconstruction system.}
	%	\vspace{-5mm}
	%	\end{figure}
	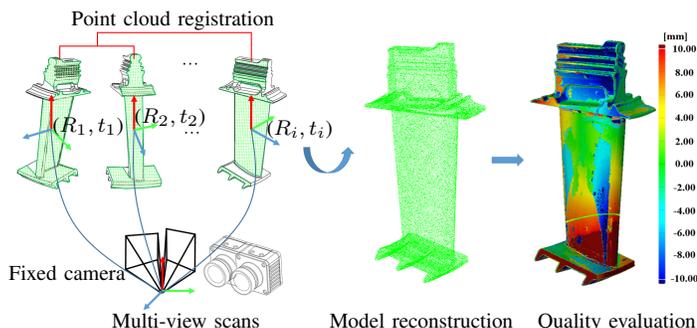
\begin{figure}[t!] 
		%		 \begin{tikzpicture}[inner sep=0pt,outer sep=0pt]
		%		\node[anchor=south west] (A) at (0in,0in)
		%		\center{\includegraphics[width=0.99\textwidth,trim=1.25in
		%			0.8in 1.3in 0.8in]{fig511.png}}
		%		
		%		\node[anchor=south,rotate=90,yshift=2pt] at (A.west)
		%		{\footnotesize Transformation deviation $d_{T}$};
		%		   \end{tikzpicture}
		%{\scalefont{2}
		\centering
		\begin{tikzpicture}[inner sep=0pt,outer sep=0pt]
		\node[anchor=south west] (A) at (0in,0in)
		{\includegraphics[width=.5\textwidth,clip=false]{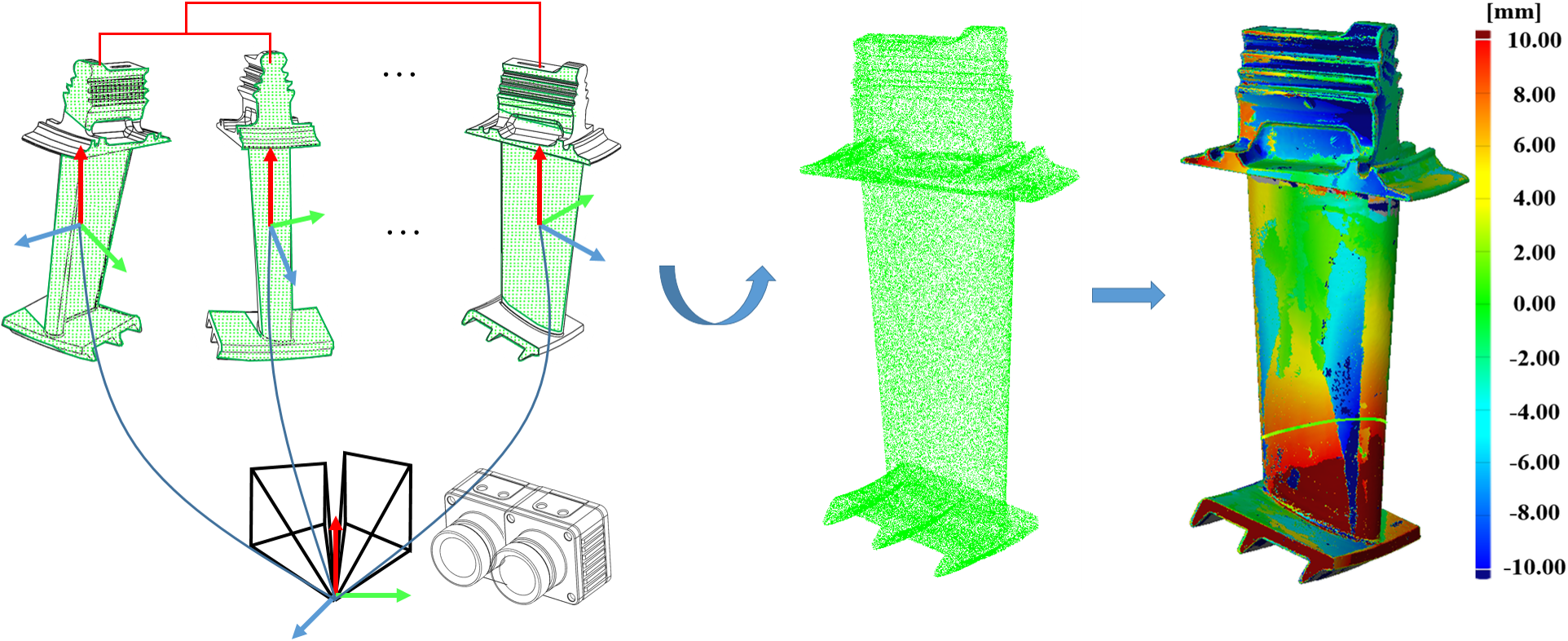}};
		%		   \node[anchor=south,rotate=90,yshift=2pt] at (A.west)
		%		   {\footnotesize Transformation deviation $d_{T}$};
    	\node[anchor=north,xshift=-65pt,yshift=-1pt] at (A.south)
    	{\footnotesize Multi-view scans};
    	\node[anchor=north,xshift=24pt,yshift=-1pt] at (A.south)
    	{\footnotesize Model reconstruction};
    	\node[anchor=north,xshift=98pt,yshift=-1pt] at (A.south)
    	{\footnotesize Quality evaluation};
    	\node[anchor=north,xshift=-102pt,yshift=+73pt] at (A.south)
    	{\footnotesize ($R_1,t_1$)};
    	\node[anchor=north,xshift=-71pt,yshift=+76.5pt] at (A.south)
    	{\footnotesize ($R_2,t_2$)};
    	\node[anchor=north,xshift=-22pt,yshift=+72pt] at (A.south)
    	{\footnotesize ($R_i,t_i$)};
	
    	\node[anchor=north,xshift=-70pt,yshift=+113.5pt] at (A.south)
    	{\footnotesize Point cloud registration};
    	\node[anchor=north,xshift=-110pt,yshift=+17pt] at (A.south)
    	{\footnotesize Fixed camera};
%		\node[anchor=north,xshift=75pt,yshift=-2pt] at (A.south)
%		{\footnotesize (d)};
		
		\end{tikzpicture}
%	}

		\vspace{-3mm}
		\caption{\label{fig:pd} Schematic diagram of the reconstruction system. The fixed camera scans multiple views of the blade with transformations. The green areas denote the visible parts of the camera view, which are then reconstructed to a complete model for quality evaluation with the standard model.}
	\end{figure}
	
	Over the past few decades, numerous methods have been developed for point cloud registration. The iterative closest point (ICP) algorithm~\cite{icp1,icp2} has been widely used for the rigid registration of 3D point clouds due to its simplicity and performance. It is intuitive and easy to implement in practice. ICP first establishes correspondences between the nearest points of the two point clouds and then minimizes the $\ell_2$ distance between the corresponding point pairs. However, the ICP algorithm is well known to be susceptible to the local minimum problem due to its assumption that the set of nearest points in the current iteration will be better correspondences than those in the last iteration.
	This assumption can easily fail when the point cloud data are contaminated with noise or have missing regions of overlap and does not guarantee a globally optimal registration. 
	
	To address the local minimum problem, Chetverikov \textit{et al.} proposed the trimmed ICP algorithm~\cite{Trimmedicp}, which enables the application of ICP to point clouds with partial overlap. Nonetheless, the original ICP and all its variants are still sensitive to noise and necessitate good initialization of the registration. Furthermore, several Gaussian mixture model (GMM)~\cite{GMM1}~\cite{CPD}~\cite{BCPD}-based methods have been proposed to seek a more robust solution. Jian \textit{et al.}~\cite{GMM1} represented the point clouds as GMMs and minimized the Kullback-Leibler divergence of two GMMs. Myronenko \textit{et al.}~\cite{CPD} regarded registration as a probability estimation problem by representing one cloud as centroids of a GMM. Hirose \textit{et al.}~\cite{BCPD} incorporated Bayesian theory into a probability model. These GMM-based methods show more robust performances than ICP but still tend to become trapped in a local minimum and usually have high time complexity. Yang \textit{et al.}~\cite{go-icp} employed a branch-and-bound scheme to search for the optimal transformation.
	A few heuristic methods~\cite{pf}~\cite{sa} have also been presented for alleviating the local minimum problem. Another strategy~\cite{lei2017fast} is to use coarse alignment with other methods, such as feature matching, to achieve good initialization. 
	However, feature-based methods are not always reliable and do not guarantee a globally optimal transformation. More critically, due to the $\ell_2$-norm least squares function~\cite{mactavish2015all}, when minimization is applied, the optimizer runs even after reaching the global minimum, especially when dealing with contaminated real point cloud data. A small number of outliers may adversely affect the validity of the results. 
	
	One solution is to employ a more robust error function instead of the sensitive $\ell_2$-norm least squares function. Wang~\textit{et al.}~\cite{ef1} formulated the misalignment problem as a quadratic integer program (minMaxQIP). Xie~\textit{et al.}~\cite{ef2} employed variance-minimization matching to replace the $\ell_2$ optimization.
	\revise{PE minimization is applied by several methods~[22]-[25] in point cloud registration. The registration is regarded as the potential energy decrease of the rigid swarms of particles with masses by Golyanik~\textit{et al.}~\cite{gaa1}. Ali~\textit{et al.}~\cite{gaa2} considered the gravitational field in locally multiply linked patches and updated the alignment parameters with unconstrained displacement fields. The more robust gravitational potential energy method BHRGA~\cite{gaa5} was proposed by Golyanik~\textit{et al.} for accelerating the computation of the potentials using a Barnes-Hut Tree. BHRGA formulates a functional gravitational potential energy (GPE) and minimizes the sum of squared residuals that are related to the $\ell_2$ loss. Yang~\textit{et al.}~\cite{gaa3} constructed a dynamical system for updating registration parameters using the Lyapunov theory. Despite their application value for PE optimization in registration problems, these methods are still based on the $\ell_2$ loss function, which would bias the registration by the perturbations. In addition, the motion that is determined by the dissipation term is often vague for the optimal convergence guarantee of gravitational approaches with altered physics.} Several methods have been proposed for addressing outliers~\cite{Trimmedicp}~\cite{OT1}~\cite{OT2}~\cite{OT3} and for consensus maximization (point pair matching maximum) between the point clouds~\cite{cmax1}~\cite{cmax2}~\cite{cmax3}. %~\cite{gaa1}~\cite{gaa2}~\cite{gaa5}~\cite{gaa3}
	Fitzgibbon \textit{et al.}~\cite{OT1} employed a standard iterative nonlinear optimizer (LM algorithm) to replace the closed-form $\ell_2$ minimization part of the ICP. Granger~\textit{et al.}~\cite{EMICP} proposed an expectation-maximization method (EMICP) for identifying and rejecting outliers. Rusinkiewicz~\textit{et al.}~\cite{OT2} assigned different weights to address outliers. 
	However, these methods are based on heuristics and incur additional computational costs. This is unacceptable for the numerous measured points in industrial measurement.
	
	Therefore, to perform precise and efficient blade reconstruction, a novel method named minimum potential energy (MPE) is developed. \revise{Different from the previously established PE-based registration method, this method optimize the proposed negative full inverse (NFI) criterion instead of the $\ell_2$ loss.} The proposed method shows optimal registration results, which are guaranteed by the shapes of the objects. The shape that is indicated by the majority of inlier points determines the results of the optimal registration and alleviates the influence of outliers and measurement noise. In summary, our contributions are as follows:
	
	\begin{enumerate}
		\item A novel \revise{NFI} criterion loss function is proposed to alleviate the influence of outliers and noise. \revise{The proposed loss function makes this method more tolerant to perturbations than methods that are influenced by the $\ell_2$ loss function.} Notably, the pairwise correspondence is not necessary for the registration guided by the NFI criterion.
		\item The classic point cloud registration problem is reformulated as an N-body simulation minimum potential energy optimization problem. \revise{We straightforwardly formulate the point cloud registration from the proposed NFI loss function for MPE optimization in (\ref{eq5}). Based on this formulation, the NFI function is optimized by the proposed motion control scheme to approach the optimal registration.}
		\item The proposed minimum potential energy algorithm is tailored for NFI criterion optimization. A physical system is constructed to facilitate the MPE registration process, which is guided by the rotational torque and gravitational vector. \revise{The motion in each iteration is reasonable for the PE minimization toward the global optimum.} We experimentally demonstrate the superior performance of our method in free-form workpiece reconstruction.
		
	\end{enumerate}

%	\begin{figure}[t!] 
%		\center{\includegraphics[width=0.5\textwidth]{problem_definition.png}}
%		\vspace{-5mm}
%		\caption{\label{fig:pd} The schematic diagram of \revise{the} reconstruction system. The fixed camera scans the different \revise{views} of %\revise{the} blade via a \revise{pose-controlled} manipulator. \revise{The} blue area denotes the visible part of \revise{the} camera view.
%		}
%		\vspace{-3mm}
%	\end{figure}

	\section{\label{ii}Problem formulation}
	Freeform surface reconstruction is the problem of finding the best transformation parameters between $K$ sets that are acquired by scanning different views and reconstructing the whole model. We consider two view scans $\mathbf{X}$ and $\mathbf{Y}$ as an example. Let $\mathbf{X}=\begin{Bmatrix}\boldsymbol{x}_1,\boldsymbol{x}_2,......,\boldsymbol{x}_N\end{Bmatrix}$ represent the template set and $\mathbf{Y}=\begin{Bmatrix}\boldsymbol{y}_1,\boldsymbol{y}_2,......,\boldsymbol{y}_M\end{Bmatrix}$ the reference set, where $D$ denotes the dimension of the data ($D=3$ in this paper) and $M,N$ are the sizes of the point sets. We denote the putative correspondences as $(\boldsymbol{x}_i, \boldsymbol{y}_i)$, where $\boldsymbol{x}_i\subset\mathbf{X}$ and $\boldsymbol{y}_i\subset\mathbf{Y}$. The optimal transformation can be expressed as a tuple $(\textit{\textbf{R}}_{k},\textit{\textbf{t}}_{k})$, where $\textit{\textbf{R}}_{k}\in SO(3)$ and $\textit{\textbf{t}}_{k}\in {\revise{\mathbb{R}}}^{3\times1}$. Hence, the former can be aligned with the latter as follows:
	\begin{equation}\label{eq1}\boldsymbol{y}_i=\textit{\textbf{R}}_{k}\boldsymbol{x}_i+\textit{\textbf{t}}_{k}+\textit{\textbf{o}}_{i}+\mathbf{\epsilon}_i, \end{equation} %%??????
	We use $\textit{\textbf{o}}_{i}$ to model the outliers and $\mathbf{\epsilon}_i$ to model the measurement noise. When we regard the correspondence pair $(\boldsymbol{x}_i, \boldsymbol{y}_i)$ as an inlier, the vector $\textit{\textbf{o}}_{i}$ remains a zero vector, which indicates that there is no error of this inlier correspondence pair under the optimal transformation, or an arbitrary vector if the correspondence involves outliers. The measurement noise $\mathbf{\epsilon}_i$ is set to model the small perturbation of the scanning process. In other words, the acquired correspondence pair deviate and are hard to align exactly in practice. The measurement noise $\mathbf{\epsilon}_i$ is intrinsic to the 3D scanning process and, hence, unavoidable. Specifically, $\boldsymbol{y}_i$ corresponds to a transformation $(\textit{\textbf{R}}_{k},\textit{\textbf{t}}_{k})$ of $\boldsymbol{x}_i$ (plus the little unknown perturbation $\mathbf{\epsilon}_i$) if $(\boldsymbol{x}_i, \boldsymbol{y}_i)$ is an inlier correspondence, whereas $\boldsymbol{y}_i$ is an arbitrary point when it belongs to be an outlier correspondence.%%????????????,????

	The popular ICP algorithm performs optimization by alternately applying the following two functions:
	\begin{equation}\label{eq4} j^*=\arg\min  \left \| \boldsymbol{y}_j-(\textit{\textbf{R}}_{k}\boldsymbol{x}_i+\textit{\textbf{t}}_{k}) \right \|,\end{equation}
	\begin{equation}\label{eq3} E(\textit{\textbf{R}},\textit{\textbf{t}})=\sum_{i=1}^{N}e_i(\textit{\textbf{R}},\textit{\textbf{t}})^2=\sum_{i=1}^{N} \left \| \boldsymbol{y}_{j^*}-(\textit{\textbf{R}}_{k}\boldsymbol{x}_i+\textit{\textbf{t}}_{k}) \right \|^2, \end{equation} 
	where $y_{j^*}$ and $\boldsymbol{x}_i$ denote the optimal corresponding point pair. Equation~\ref{eq3} implements the transformation estimation, and (\ref{eq4}) matches the closest points. 	
	To obtain the optimal registration, the parameter $\textit{\textbf{o}}_{i}$ should be a zero coordinate vector by searching for the inliers and discarding outlier correspondences. Subsequently, the 3D transformation is estimated to minimize the $\ell_2$ error of putative inlier correspondences with the measurement $\mathbf{\epsilon}_i$. %%%bi-ai=epsilon_i;
	Despite its many desirable properties, including simplicity, ICP implicitly requires full overlap between the point clouds, which is a rare situation in practice. Applying the $\ell_2$ metric in~(\ref{eq3}) and taking the nearest neighbor points in~(\ref{eq4}) as the corresponding points makes the ICP algorithm susceptible to becoming trapped in a local minimum.
	
\begin{figure}[t!] 
		\center{\includegraphics[width=0.5\textwidth]{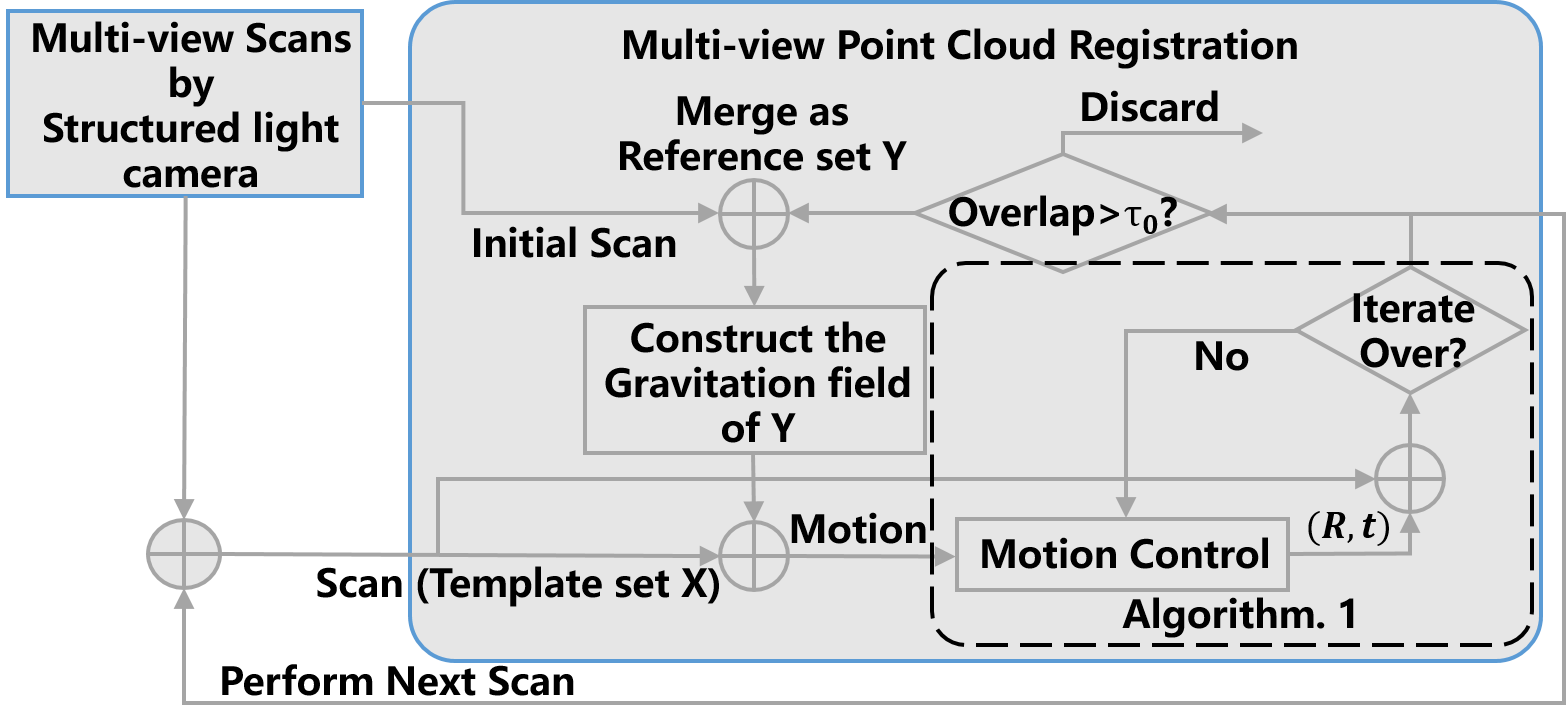}}
		\vspace{-5mm}
		\caption{\label{fig:zz} Framework of the proposed MPE method. 
		} 
	\end{figure}

	As shown in Fig.~\ref{fig:pd}, the pose of a blade is transformed to obtain different view scans with a fixed camera. The green area denotes a part of the blade that has been scanned. Then, the set of scans is reconstructed to the completed model by point cloud registration. Similar to the two-point set registration that is discussed above, we match $K$ sets to reconstruct the completed model. The framework of the proposed multiview registration method is illustrated in Fig.~\ref{fig:zz}. We decompose the multiview registration into two point set registrations. Once the registration of the current two scenes is finished, a decision is made as to whether to merge the aligned points as the new reference set when the overlap of current point sets is more than $\tau_0$. Otherwise, the current template set is assumed to contain unqualified points and, hence, is discarded.

	%?????,????????

	\section{Approach}
	In this section, we consider two view scan registration as an example to demonstrate how we establish a correspondence between views. As defined in Section~\ref{ii}, the point cloud registration problem generally involves two D-dimensional point sets: a reference set $\mathbf{Y}$ and a template set $\mathbf{X}$. 
	First, to overcome the local minimum problem of current methods, we propose the NFI criterion (\ref{eq33}) in Section~\ref{iiia} to perform optimization to achieve a global minimum. Next, based on the proposed negative full inverse (NFI) criterion, we assume in Section~\ref{iiic} that the minimum potential energy position of the system is equivalent to the global optimum. Hence, the optimal NFI function can be solved by approaching the minimum potential energy of the whole system.
	In Section~\ref{iiid}, we present a motion control procedure that reduces the potential energy of the whole system to a minimum in the constructed physical system. In summary, we construct a physical movement algorithm for optimizing the proposed NFI criterion instead of the sensitive $L_2$ criterion.

	\subsection{\label{iiia} Negative Full Inverse (NFI) Criterion}%negative full inverse metric.
	To overcome the local minimum problem, a global registration method must have two properties. First, the local matching process (e.g., based on nearest points) must be replaced with a process that is influenced globally. 
	Second, different weights should be allocated to the points to alleviate the influence of outliers. 
	It is easy to see from (~\ref{eq3}) that most current registrations are optimized by minimizing the $L_2$ distance function $E(\textit{\textbf{R}},\textit{\textbf{t}})$. However, it is well known that the $L_2$-norm least-squares metric is not robust, as it is sensitive to outliers. To make matters worse in the specific case of point set registration based on $L_2$-norm optimization, incorrect matching pairs are assigned higher weights since their Euclidean distance is very large. Such pairs should be assigned lower weights. 
	
	Hence, we apply the inverse proportion function to calculate the pairwise distance against the $\ell_2$ metric for assigning different weights in (\ref{eq3}) and consider a globally full pairwise connection against nearest neighbor points, as expressed in (\ref{eq4}). Furthermore, a negative term is added to reverse the optimization direction to minimization. The inverse proportion function, full pairwise connection, and negative term are denoted as I, F, and N, respectively. We complete the alignment by minimizing the following criterion, which is named the negative full inverse (NFI) function:
	\begin{equation}
	\begin{aligned}
	\label{eq33} E_I(\textit{\textbf{R}},\textit{\textbf{t}})&=NFI\{  e_{ij}(\textit{\textbf{R}},\textit{\textbf{t}}) \} \\
	&=-\sum_{i=1}^{N}\sum_{j=1}^{M} \frac{1}{\left \| \boldsymbol{y}_j-(\textit{\textbf{R}}\boldsymbol{x}_i+\textit{\textbf{t}}) \right \|\revise{+\varepsilon^2}},
	\end{aligned}
	\end{equation} 
	where $e_{ij}(\textit{\textbf{R}},\textit{\textbf{t}})$ denotes the per-point residual error of $\boldsymbol{x}_i$ and $\boldsymbol{y}_j$ and $\varepsilon^2$ represents a constant that ensures a lower bound of a single correspondence pair. We introduce the nonzero real number $\varepsilon^2$ for the motion control to produce an infinitesimal minimum potential energy of a single correspondence pair. It prevents the movement of the template set from being frozen by the infinite value of a single correspondence pair and, thus, becoming trapped in local minimum (singularity), which is akin to a black hole. 
	There are two advantages to using the outlier per-point residual $\textit{\textbf{o}}_{i}$ and measurement noise $\mathbf{\epsilon}_i$ in (\ref{eq1}) with the NFI function: $1)$ The per-point residual $\textit{\textbf{o}}_{i}$ of outliers is assigned lower weight by the NFI function as the residual error increases, whereas the $\ell_2$ function gives more attention to outliers. $2)$ Measurement noise $\mathbf{\epsilon}_i$ is considered globally. The scheme considers the global optimum with the small perturbation $\mathbf{\epsilon}_i$. In contrast, the ICP method only considers the nearest neighbor of each point, which is one of the main reasons that it is vulnerable to the local minimum problem.
	In other words, given $\textit{\textbf{R}}\in SO(3)$ and $\textit{\textbf{t}}\in {\mathbb{R}}^3$, the NFI criterion assigns more weight to inlier correspondence and less weight to outliers, instead of only considering the nearest neighbors and giving no attention to points with long distances; hence, it improves the robustness of the proposed alignment method.
	
	The loss function of a method is a significant guideline of optimization. The traditional $L_2$ distance method in ICP uses a nonlinear loss function $ min(\left \| \boldsymbol{y}_{i}-(\textit{\textbf{R}}\boldsymbol{x}_i+\textit{\textbf{t}}) \right \|^2) $ of the point pairs to assign weights at different distances. This loss function generally performs satisfactorily but fails in the presence of measurement noise and outliers. Perturbations, especially outliers with longer distances, are assigned quadratic weights and bias the registration.
	Our proposed NFI criterion easily ignores noise and assigns lower weights to points that are~\revise{farther}. \revise{In conclusion, the NFI criterion optimizes the registration by giving more weight to inliers. The inlier registration is regarded as a maximum common set between two aligned sets and is located at an optimum.} 
	
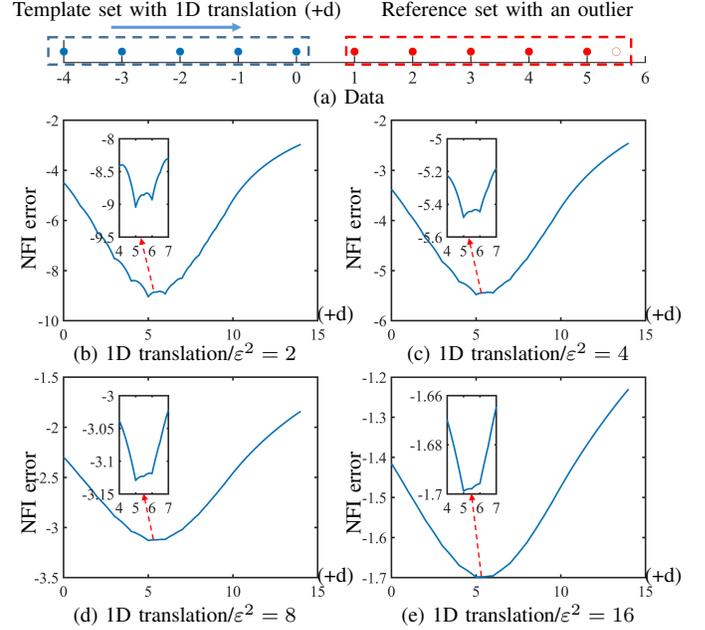
\begin{figure}[t!] 
		%\center{\includegraphics[width=0.45\textwidth]{fig111.eps}}
		%图例还没
		\begin{tikzpicture}[inner sep=0pt,outer sep=0pt]
		\node[anchor=south west] (A) at (0in,0in)
		{\includegraphics[width=.48\textwidth,clip=false]{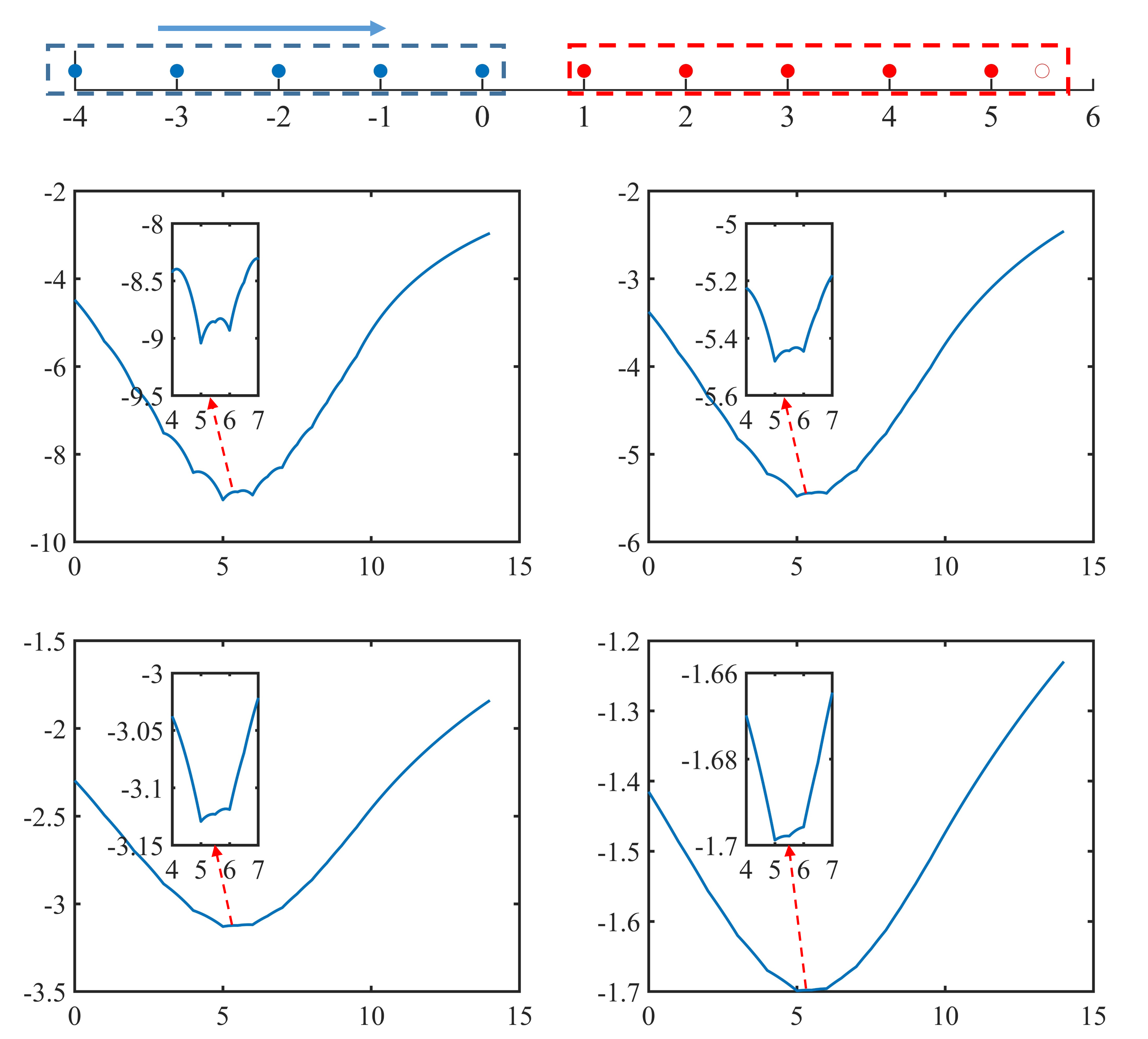}};
		%		   \node[anchor=south,rotate=90,yshift=2pt] at (A.west)
		%		   {\footnotesize Transformation deviation $d_{T}$};
		%	\node[anchor=north,xshift=0pt,yshift=178-5pt] at (A.south)
		%	{\footnotesize (a) Gaussian Noise\%};
		%	\node[anchor=north,xshift=-8pt,yshift=93-9pt,rotate=90] at (A.west)
		%	{\footnotesize Success Rate};
		%	
							\node[anchor=north,xshift=-65pt,yshift=240-12pt] at (A.south)
		{\footnotesize Template set with 1D translation (+d)};
							\node[anchor=north,xshift=60pt,yshift=240-12pt] at (A.south)
		{\footnotesize Reference set with an outlier};
		
						\node[anchor=north,xshift=0pt,yshift=210-15pt] at (A.south)
		{\footnotesize (a) Data};
		
					\node[anchor=north,xshift=-62pt,yshift=99pt] at (A.south)
				{\footnotesize (b) 1D translation/$\varepsilon^2=2$};
								\node[anchor=north,xshift=64pt,yshift=99pt] at (A.south)
				{\footnotesize (c) 1D translation/$\varepsilon^2=4$};
				
								\node[anchor=north,xshift=-62pt,yshift=0pt] at (A.south)
				{\footnotesize (d) 1D translation/$\varepsilon^2=8$};
								\node[anchor=north,xshift=64pt,yshift=0pt] at (A.south)
				{\footnotesize (e) 1D translation/$\varepsilon^2=16$};
				
				\node[anchor=north,xshift=-62+57pt,yshift=99+14pt] at (A.south)
				{\footnotesize (+d)};
				\node[anchor=north,xshift=64+55pt,yshift=99+14pt] at (A.south)
				{\footnotesize (+d)};
				
				\node[anchor=north,xshift=-62+57pt,yshift=14pt] at (A.south)
				{\footnotesize (+d)};
				\node[anchor=north,xshift=64+55pt,yshift=14pt] at (A.south)
				{\footnotesize (+d)};

							\node[anchor=north,xshift=0pt,yshift=30pt,rotate=90] at (A.west)
				{\footnotesize NFI error};
							\node[anchor=north,xshift=0pt,yshift=-70pt,rotate=90] at (A.west)
				{\footnotesize NFI error};
							\node[anchor=north,xshift=123pt,yshift=30pt,rotate=90] at (A.west)
				{\footnotesize NFI error};
							\node[anchor=north,xshift=123pt,yshift=-70pt,rotate=90] at (A.west)
				{\footnotesize NFI error};
		%	
		%	\node[anchor=north,xshift=0pt,yshift=90-5pt] at (A.south)
		%	{\footnotesize (b) Incomplete\%};		
		
		%	
		%	\node[anchor=north,xshift=0pt,yshift=0-3pt] at (A.south)
		%	{\footnotesize (c) Outliers Number(\#pts)};
		%	\node[anchor=north,xshift=-8pt,yshift=-93pt,rotate=90] at (A.west)
		%	{\footnotesize Success Rate};

		%		\node[anchor=north,xshift=75pt,yshift=92pt] at (A.south)
		%		{\footnotesize (b)};
		%		\node[anchor=north,xshift=-186pt,yshift=-2pt] at (A.south)
		%		{\footnotesize (c)};
		%		\node[anchor=north,xshift=75pt,yshift=-2pt] at (A.south)
		%		{\footnotesize (d)};
		
		\end{tikzpicture}

		\vspace{-3mm}
		\caption{\label{fig:P4} \revise{One-dimensional illustration of the registration error with the NFI metrics. (a) 1D data, the blue template set and the red reference set. The hollow circle in the reference set denotes an outlier. (b)-(e) The curve of the NFI loss function with the 1D translation of the template set.} }
	\end{figure}

	\revise{We present a series of examples of lower dimension to better visualize the advantages of the NFI function. The impact of the parameter $\varepsilon^2$ selection is shown in Fig.~\ref{fig:P4}. It shows a 1D illustration of the registration error with the NFI metrics by the 1D translation of the template set. We present the 1D registration using an extra outlier with a deviation $0.5$ from the nearest real match in the reference set. An enlarged view around the optimum is also presented. It is obvious that a local minimum is attained when the perturbation is close to the real match. However, the smoothness of the error curve can be adjusted using suitable parameters. $\varepsilon^2=4$ results in a smoother peak and trough on the error curve. When the value is increased to $8$, the error function becomes a convex function. We continue to improve the parameter and show that parameter adjustment is feasible.}
	
\begin{figure*}[t!] 
		%\center{\includegraphics[width=0.45\textwidth]{fig111.eps}}
		%图例还没
		\begin{tikzpicture}[inner sep=0pt,outer sep=0pt]
		\node[anchor=south west] (A) at (0in,0in)
		{\includegraphics[width=.99\textwidth,clip=false]{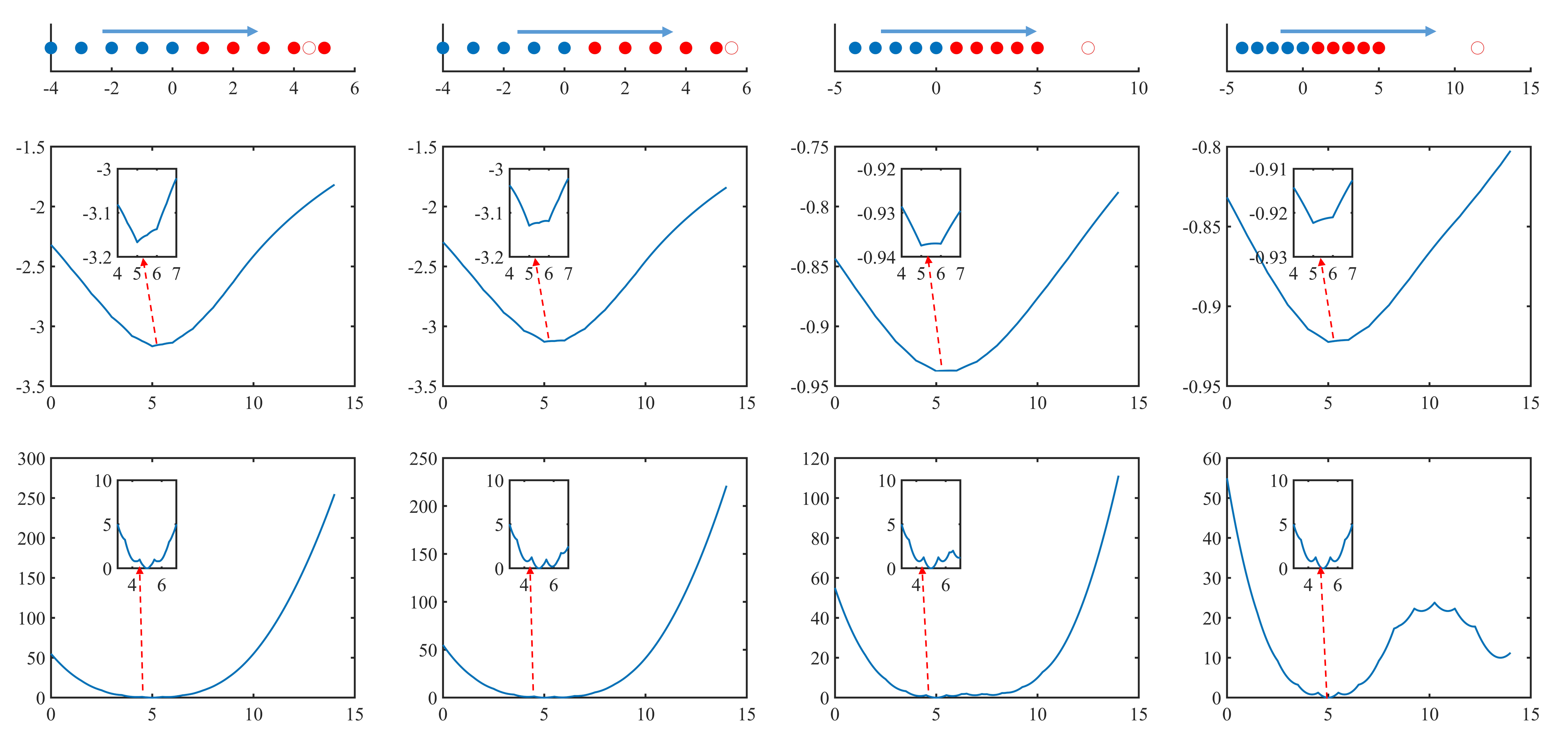}};
		%		   \node[anchor=south,rotate=90,yshift=2pt] at (A.west)
		%		   {\footnotesize Transformation deviation $d_{T}$};
		%	\node[anchor=north,xshift=0pt,yshift=178-5pt] at (A.south)
		%	{\footnotesize (a) Gaussian Noise\%};
		%	\node[anchor=north,xshift=-8pt,yshift=93-9pt,rotate=90] at (A.west)
		%	{\footnotesize Success Rate};
		%	
	\node[anchor=north,xshift=-199pt,yshift=240-4pt] at (A.south)
{\footnotesize 1D translation (+d)};
\node[anchor=north,xshift=-64pt,yshift=240-4pt] at (A.south)
{\footnotesize 1D translation (+d)};
\node[anchor=north,xshift=55pt,yshift=240-4pt] at (A.south)
{\footnotesize 1D translation (+d)};
\node[anchor=north,xshift=185pt,yshift=240-4pt] at (A.south)
{\footnotesize 1D translation (+d)};
	
	\node[anchor=north,xshift=-195pt,yshift=210-8pt] at (A.south)
	{\footnotesize (a) Data/outlier$=4.5$};
		\node[anchor=north,xshift=-65pt,yshift=210-8pt] at (A.south)
	{\footnotesize (b) Data/outlier$=5.5$};
		\node[anchor=north,xshift=65pt,yshift=210-8pt] at (A.south)
	{\footnotesize (c) Data/outlier$=7.5$};
		\node[anchor=north,xshift=200pt,yshift=210-8pt] at (A.south)
	{\footnotesize (d) Data/outlier$=11.5$};
	
	\node[anchor=north,xshift=-195pt,yshift=100pt] at (A.south)
	{\footnotesize (e) 1D translation/outlier$=4.5$};
	\node[anchor=north,xshift=-65pt,yshift=100pt] at (A.south)
	{\footnotesize (f) 1D translation/outlier$=5.5$};
		\node[anchor=north,xshift=65pt,yshift=100pt] at (A.south)
	{\footnotesize (g) 1D translation/outlier$=7.5$};
		\node[anchor=north,xshift=200pt,yshift=100pt] at (A.south)
	{\footnotesize (h) 1D translation/outlier$=11.5$};
	
	\node[anchor=north,xshift=-195pt,yshift=-2pt] at (A.south)
	{\footnotesize (i) 1D translation/outlier$=4.5$};
	\node[anchor=north,xshift=-65pt,yshift=-2pt] at (A.south)
	{\footnotesize (j) 1D translation/outlier$=5.5$};
		\node[anchor=north,xshift=65pt,yshift=-2pt] at (A.south)
	{\footnotesize (k) 1D translation/outlier$=7.5$};
		\node[anchor=north,xshift=200pt,yshift=-2pt] at (A.south)
	{\footnotesize (l) 1D translation/outlier$=11.5$};
	
		\node[anchor=north,xshift=-195+62pt,yshift=100+17pt] at (A.south)
	{\footnotesize (+d)};
	\node[anchor=north,xshift=-65+59pt,yshift=100+17pt] at (A.south)
	{\footnotesize (+d)};
	\node[anchor=north,xshift=65+57pt,yshift=100+17pt] at (A.south)
	{\footnotesize (+d)};
	\node[anchor=north,xshift=200+50pt,yshift=100+17pt] at (A.south)
	{\footnotesize (+d)};
	
	\node[anchor=north,xshift=-195+62pt,yshift=-2+17pt] at (A.south)
	{\footnotesize (+d)};
	\node[anchor=north,xshift=-65+59pt,yshift=-2+17pt] at (A.south)
	{\footnotesize (+d)};
	\node[anchor=north,xshift=65+57pt,yshift=-2+17pt] at (A.south)
	{\footnotesize (+d)};
	\node[anchor=north,xshift=200+50pt,yshift=-2+17pt] at (A.south)
	{\footnotesize (+d)};

			\node[anchor=north,xshift=-2pt,yshift=34pt,rotate=90] at (A.west)
	{\footnotesize NFI error};
			\node[anchor=north,xshift=126pt,yshift=34pt,rotate=90] at (A.west)
{\footnotesize NFI error};
			\node[anchor=north,xshift=253pt,yshift=34pt,rotate=90] at (A.west)
{\footnotesize NFI error};
			\node[anchor=north,xshift=381pt,yshift=34pt,rotate=90] at (A.west)
{\footnotesize NFI error};

			\node[anchor=north,xshift=-2pt,yshift=-68pt,rotate=90] at (A.west)
{\footnotesize $\ell_2$ error};
			\node[anchor=north,xshift=126pt,yshift=-68pt,rotate=90] at (A.west)
{\footnotesize $\ell_2$ error};
			\node[anchor=north,xshift=253pt,yshift=-68pt,rotate=90] at (A.west)
{\footnotesize $\ell_2$ error};
			\node[anchor=north,xshift=381pt,yshift=-68pt,rotate=90] at (A.west)
{\footnotesize $\ell_2$ error};
		\end{tikzpicture}

		\vspace{-3mm}
		\caption{\label{fig:P3} \revise{One-dimensional illustration of the registration error with the traditional $\ell_2$ error and the NFI metrics with a value of $\varepsilon^2$. Each column represents the loss function map with an outlier in the reference set. The first row (a)-(d) presents the data with four different outlier values. Each 1D data set consists of a blue template set and a red reference set. The hollow circle in the reference set denotes an outlier. The second row (e)-(h) presents the corresponding NFI error, and the last row presents the $\ell_2$ error. For a better illustration of the difference, we enlarge the local map in an extra window around the optimal translation $+5$.} }
	\end{figure*}
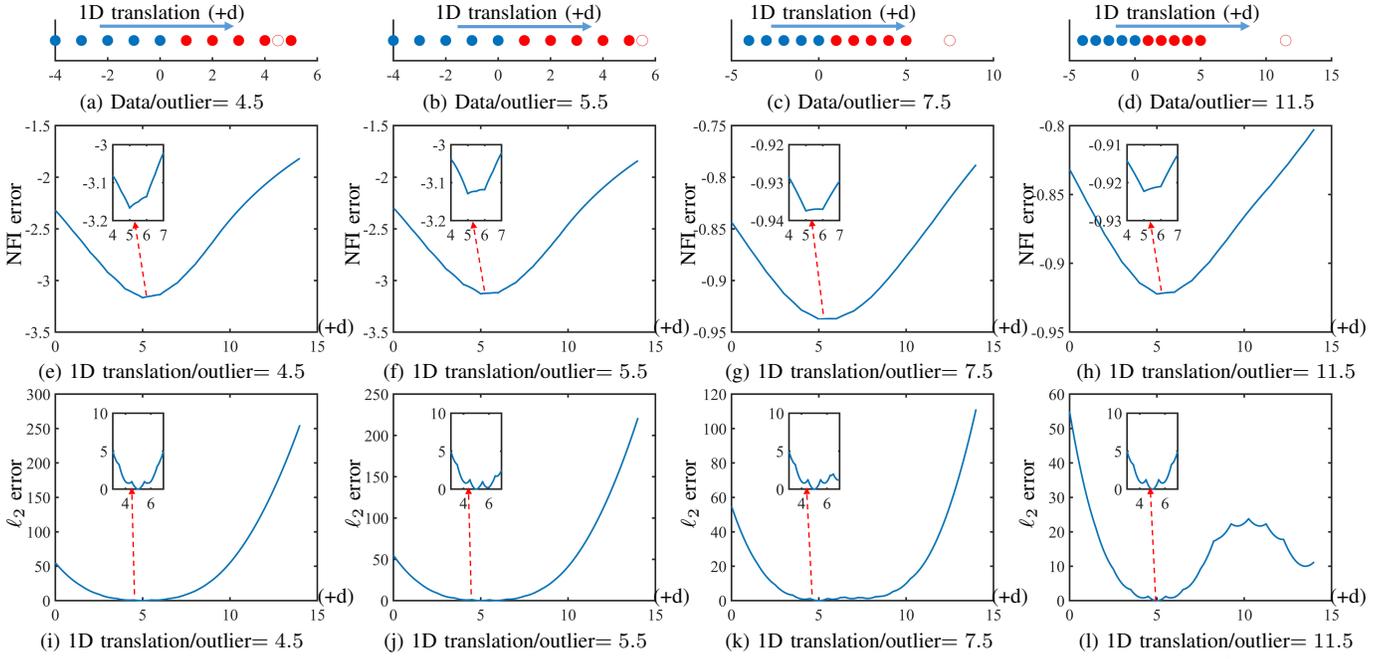

	\revise{Like Fig.~\ref{fig:P4}, we compare the NFI function and $\ell_2$ metric as the distance of the outlier increases in Fig.~\ref{fig:P3}. Each column presents a comparison of the error map under different outlier distances. The four distances model different types of outliers from near to far and are shown in Fig.~\ref{fig:P3}(a)-(d). It is easy to see that NFI error functions are convex and consist of one single optimum and no other local minimum, whereas the error functions contain multiple local minima when applying the $\ell_2$ loss. In particular, compared to the local minima around the optima, the $\ell_2$ loss function shows a sharper and wider local minimum around $\text{translation}+14$, which usually has a higher possibility of trapping the optimization in the wrong convergence.}
	
	\revise{With perturbations, we can see that the NFI criterion is more applicable for registration. The inliers form a maximum common set to ensure convex optimization and are assigned higher weight by the NFI criterion. These example shows that our NFI criterion can address the various perturbation distances. This can be explained as follows: There is no correspondence for the perturbation alignment. However, perturbation for alignment with one other point would bias the majority of the inlier correspondences in the registration, which would result in a substantial penalty by the NFI function. The number of the inlier correspondences is much greater than the number of perturbations in practical blade measurement. }
	
	%We discuss the advantages of NFI criterion compared to %$\ell_2$% loss. We also present the setting when dealing with the different distance of the perturbation. The limitation is discussed in the end.  距离越大，影响越大 \revise{We employ the NFI criterion to solve the determinstic problem that the $\ell_2$ loss method usually fails when outliers baised registration result.}

		\begin{figure}[t!] 
		%\center{\includegraphics[width=0.45\textwidth]{fig111.eps}}
		%图例还没
		\begin{tikzpicture}[inner sep=0pt,outer sep=0pt]
		\node[anchor=south west] (A) at (0in,0in)
		{\includegraphics[width=.48\textwidth,clip=false]{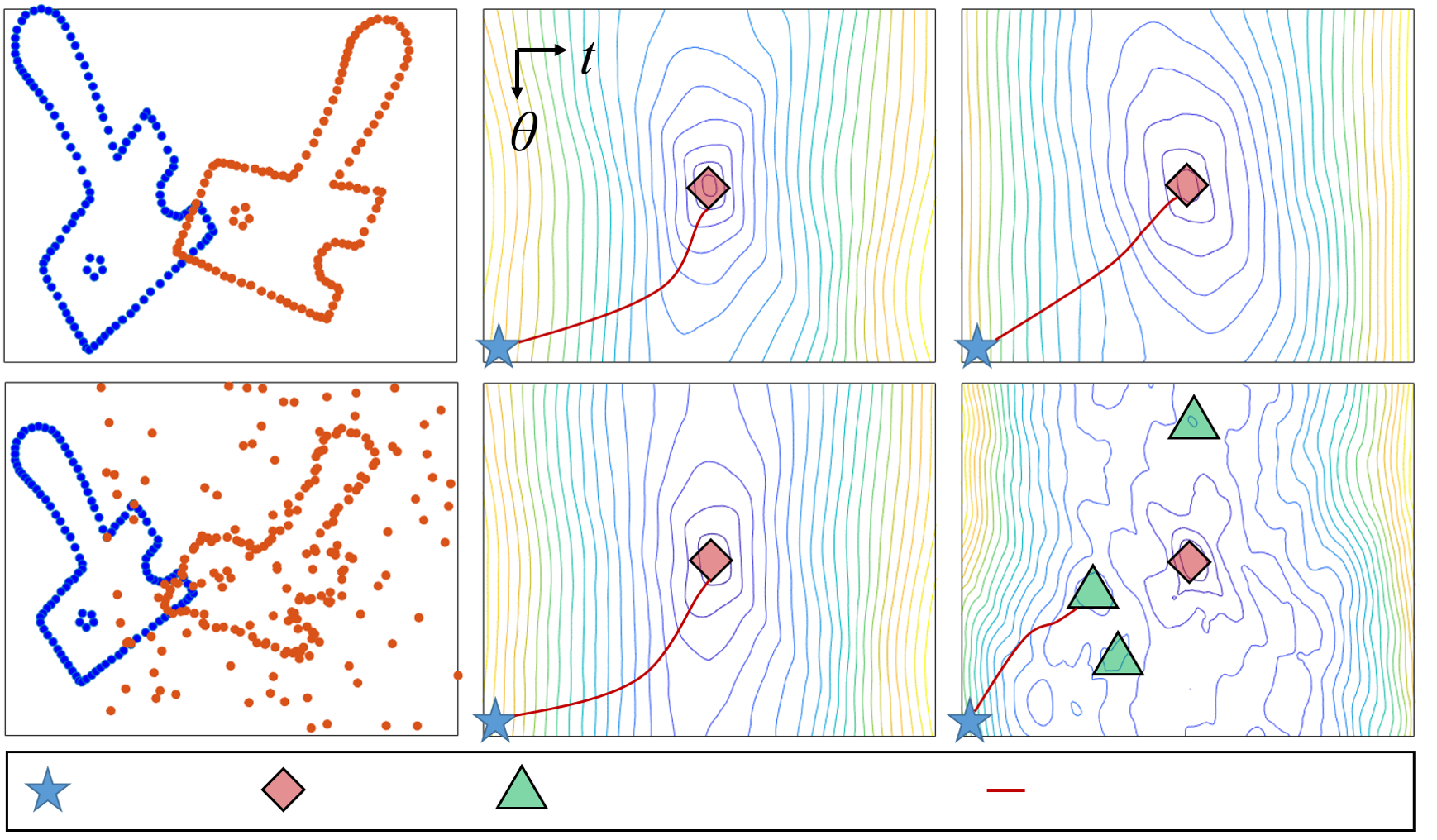}};
		%		   \node[anchor=south,rotate=90,yshift=2pt] at (A.west)
		%		   {\footnotesize Transformation deviation $d_{T}$};
		%	\node[anchor=north,xshift=0pt,yshift=178-5pt] at (A.south)
		%	{\footnotesize (a) Gaussian Noise\%};
		%	\node[anchor=north,xshift=-8pt,yshift=93-9pt,rotate=90] at (A.west)
		%	{\footnotesize Success Rate};
		%	
		%%			\node[anchor=north,xshift=-88pt,yshift=187-9pt] at (A.south)
		%		{\footnotesize (a)};
		%	%		\node[anchor=north,xshift=-27pt,yshift=187-9pt] at (A.south)
		%	%	{\footnotesize (b)};
		%	%		\node[anchor=north,xshift=33pt,yshift=187-9pt] at (A.south)
		%	%	{\footnotesize (c)};
		%	%		\node[anchor=north,xshift=93pt,yshift=187-9pt] at (A.south)
		%	%	{\footnotesize (d)};
		%	
		%	\node[anchor=north,xshift=0pt,yshift=90-5pt] at (A.south)
		%	{\footnotesize (b) Incomplete\%};		
		\node[anchor=north,xshift=-8pt,yshift=0+40pt,rotate=90] at (A.west)
		{\footnotesize (a) Clean Data};
		\node[anchor=north,xshift=-8pt,yshift=-25pt,rotate=90] at (A.west)
		{\footnotesize (b) Noised Data};

		\node[anchor=north,xshift=-97pt,yshift=0+10.5pt] at (A.south)
		{\footnotesize  Init. Pt.};
		\node[anchor=north,xshift=-58pt,yshift=0+10.5pt] at (A.south)
		{\footnotesize GT Pt. };
		\node[anchor=north,xshift=7pt,yshift=0+10.5pt] at (A.south)
		{\footnotesize Some local minimum};
		\node[anchor=north,xshift=86pt,yshift=0+10.5pt] at (A.south)
		{\footnotesize Optimization path};

		\node[anchor=north,xshift=-80pt,yshift=0+8pt] at (A.north)
		{\footnotesize blade cross-section};
		\node[anchor=north,xshift=0pt,yshift=0+8pt] at (A.north)
		{\footnotesize NFI};
		\node[anchor=north,xshift=84pt,yshift=0+8pt] at (A.north)
		{\footnotesize $\ell_2$};
		%	
		%	\node[anchor=north,xshift=0pt,yshift=0-3pt] at (A.south)
		%	{\footnotesize (c) Outliers Number(\#pts)};
		%	\node[anchor=north,xshift=-8pt,yshift=-93pt,rotate=90] at (A.west)
		%	{\footnotesize Success Rate};

		%		\node[anchor=north,xshift=75pt,yshift=92pt] at (A.south)
		%		{\footnotesize (b)};
		%		\node[anchor=north,xshift=-186pt,yshift=-2pt] at (A.south)
		%		{\footnotesize (c)};
		%		\node[anchor=north,xshift=75pt,yshift=-2pt] at (A.south)
		%		{\footnotesize (d)};
		
		\end{tikzpicture}

		\vspace{-3mm}
		\caption{\label{fig:P2} Two-dimensional illustration of the registration error with traditional $\ell_2$ and NFI metrics. The rows from top to bottom present the data, the error contour map using the NFI metric, and using the $\ell_2$ metric. The columns present the noiseless and corrupted data registrations with the 2D cross-section point set of the blade. The relative positions (abbreviated as Pt.) of the registration are represented as a star, rhombus, and triangle. \revise{The four contour maps of the last two columns share the same coordinate system with rotation $\theta$ and translation $t$}.}
	\end{figure}
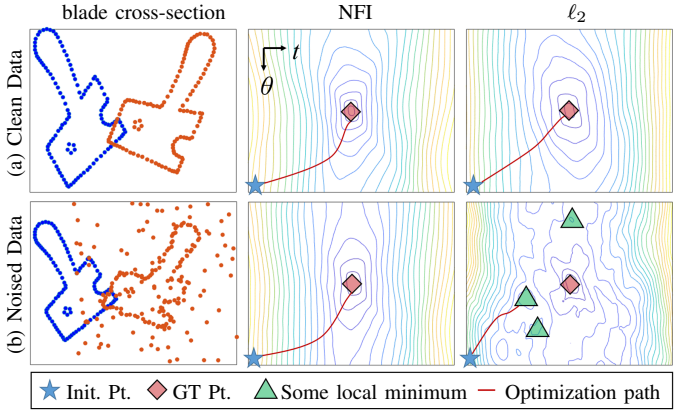
	
	\revise{Furthermore, we add a rotation dimension to the previously considered 1D translation in Fig.~\ref{fig:P2}. \revise{With the 2D map, the potential path is more vivid to show the optimization process.} It} shows a toy 2D example to illustrate the difference in registration that results from the NFI and $\ell_2$ criterion with the clean data and the noisy data in the presence of measurement noise and outliers. \revise{The four contour maps share the same horizontal and vertical axes.} The first column shows clean and noisy data images that are used to generate an error map of the NFI criterion of our MPE method and the $\ell_2$ metric of the ICP metric. Due to the requirements of the matching process for the $\ell_2$ metric, we employ nearest neighbor search for the per-point residual, just as the ICP method does.
	The two axes of the error map are the rotation $\theta_t$ and a linear translation $\textit{\textbf{t}}$ to transform the blue point set. The red rhombus is the position (location) of the ground truth (GT), and the green triangles are local minima. The red lines denote the potential paths of optimization. 
	When processing the clean data, NFI and $\ell_2$ both show an easy convergence of the error contour map due to the single optimal position. It is obvious that the optimization to GT of the NFI function is undisturbed. \revise{The noisy data consist of Gaussian noise and outliers with different distances. These perturbations bias the value of the NFI map, but the function is still convex.} However, the local minima of the error map with the $\ell_2$ function have a higher possibility of disturbing the convergence. \revise{This is the main reason that the current registration methods usually fail to obtain real blade measurements in complex environments.}
	
	The main reason for the difference is that the NFI function considers the majority of the inliers in the correspondences registration. The perturbations, especially the outliers with a larger distance to the main part of the point set are given a much lower weight in the registration process.~\revise{1) With a far distance of perturbations, the NFI function distributes much more weight to inliers. In this situation, the NFI shows great performance when dealing with outliers and noises. 2) With a near distance of perturbations, the NFI function relies on the optimal transformation of the majority of the inliers. Because there is no true correspondence of outliers in registration. The penalty to align the wrong position of the outliers is exponentially higher than to align true inliers correspondence. In this situation, the extendable $\varepsilon^2$ is adjusted to prevent the registration from being trapped by some number of the wrong correspondences.}

	\subsection{\label{iiic} Minimum potential energy registration conversion}
	%%??????
	%%????????
	In this section, we introduce a physical system into this problem to facilitate efficient registration. The system is adapted from the real world by adding a series of modifications to adjust it to the problem at hand. These modifications and assumptions are presented below:
	
	\begin{enumerate}
		\item Every point is assumed to be a particle that has a mass but no volume to avoid collision issues in the system. This ensures that the motion of the whole system is only controlled by gravity.
		\item The point cloud $\mathbf{X}$ is considered to be a rigid body. The force between its internal points corresponds to the internal force of the system and does not affect its motion. This design is due to the reconstructed point cloud being rigid and there being no deformation.
		\item The point cloud $\mathbf{X}$ resides in the constant inhomogeneous force field that is induced by the point cloud $\mathbf{Y}$. We fix the position of the point cloud $\mathbf{Y}$ and induce a constant inhomogeneous gravitational field to control the motion of $\mathbf{Y}$.
		\item The system does not follow the law of conservation of kinetic energy, namely, the system is not isolated. Moreover, each position of $\mathbf{X}$ is computed, and its distribution is discrete. Our system does not contain kinetic energy or follow the law of conservation of energy. In each iteration, we just move the point cloud $\mathbf{X}$ in the direction of decreasing the potential energy of the system. Each step is discrete and only depends on the current status.
		\item The potential energy at infinity is regarded as zero. It provides a datum of potential energy.
	\end{enumerate}

	We first propose MPE, which is a physics-inspired procedure that is based on the modified N-body simulation, for fine registration. Unlike ICP, which relies on closest point pairs, we use the gravitational field that is induced by the set of points (particles) to implement the registration.
	
	To compare the physical parameters with the registration parameters, the physical position minimum $E(\textit{\textbf{R}},\textit{\textbf{t}})$~\cite{miniPE} is the best registration position. The PE of the whole system is formulated as
	\begin{equation}
	\begin{aligned}
	\label{eq5}\phi&=\sum_{i=1}^{N}\sum_{j=1}^{M} \int^{\infty}_{r} -\frac{Gm_{i}m_{j}}{r^2} dr \\
	&=\sum_{i=1}^{N}\sum_{j=1}^{M}  (\frac{Gm_{i}m_{j}}{r})\bigg|_{r=r}^{r=\infty}\\
	&=\sum_{i=1}^{N}\sum_{j=1}^{M}  -\frac{Gm_{i}m_{j}}{r},
	\end{aligned}
	\end{equation} 
	where $r=\left \| \boldsymbol{y}_j-(\textit{\textbf{R}}\boldsymbol{x}_i+\textit{\textbf{t}}) \right \|\revise{+\varepsilon^2}$ denotes the Euclidean distance between points $(i,j)$, $m_{i},m_{j}$ represent the masses of the particles, and G denotes the gravitational constant. The PE of the system follows the fifth assumption: it is zero at infinity.
	The minimization of potential energy is derived as
	\begin{equation}
	\begin{aligned}
	\arg \min \label{eq5d}\phi&\propto \arg \min E_I(\textit{\textbf{R}},\textit{\textbf{t}}),
	\end{aligned}
	\end{equation} 
	
	Hence, we minimize the potential energy of the system to approach NFI minima and obtain the optimal alignment. According to the law of entropy increase, energy always flows from a higher level to a lower level, and the motion of the whole system is directed toward the minimum potential energy position. 
	%Quality Control Editor: Please ensure that the intended meaning has been maintained in the following edit. 

	We construct a motion control procedure of the physical system for approaching the PE minimum under the optimal NFI criterion.

	\begin{algorithm}
		\label{algorithm1}
		\SetAlgoLined
		\KwIn{Point clouds P and Q}
		\KwOut{Rotation matrix $\textit{\textbf{R}}$ and translation vector \textit{\textbf{t}}}
		init flags $F_R = 0$ and $F_t = 0$ and strides $\theta_t$ and $s_t$;\\
		Point clouds P and Q are downsampled to $P_{s}$ and $Q_{s}$;\\
		\While{$\theta_t>\varepsilon_R$ AND $s_t>\varepsilon_t$}{
			$\textit{\textbf{n}}_{p(last)}=\textit{\textbf{n}}_{p}$, $\textit{\textbf{v}}_{t(last)}=\textit{\textbf{v}}_{t}$;\\
			$F_R$ and $F_t$ are compuated with~(\ref{RR})(\ref{tt});\\
			\If{$F_R <0$}{
				$\theta_t=\frac{\theta_t}{2}$;\\
				\If{$F_t <0$}{
					$s_t=\frac{s_t}{2}$;\\
				}
			}
			Q is transformed with $\theta_t, s_t$ according to (\ref{810})(\ref{eqRf});	
		}
		The trimmed-ICP algorithm is applied to approach the fine registration;
		\caption{Minimum potential energy (MPE) algorithm for the traction force for optimal registration}
	\end{algorithm}

	\subsection{\label{iiid} Motion Control Optimization}
	In the previous section, we obtained the minimum potential energy~(\ref{eq5}), which is proportional to the loss function optimization of registration~(\ref{eq33}). 
	However, it is still difficult to obtain a mathematical solution for the MPE formulation. Therefore, we present our motion control scheme for searching for the MPE of the system in this section.

	Algorithm \ref{algorithm1} details the steps of the proposed MPE alignment. 
	In this section, we attempt to employ gravitation and obtain the direction of motion for potential energy minimization. After obtaining the direction of motion, we control the strides, including the rotation angle $\theta_t$ and translation stride $s_t$, to constrain the registration around the potential energy minimum and iteratively approach the optimum.
	
	\revise{\emph{\textbf{Per-point gravitation in $\mathbf{X}$:}} We first define the per-point gravitation $\boldsymbol{F}_{xi}$ in $\mathbf{X}$ that is attracted by $\mathbf{Y}$. It is formulated as:
	\begin{equation}
	\label{eq6} 
	\boldsymbol{F}_{xi}=-Gm_{i}\sum_{j=1}^{M}\frac{m_{j}}{ r^2}\cdot \boldsymbol{n}_{ij},
	\end{equation}
	where $\boldsymbol{n}_{ij}$ is the normalized vector that points from particle $\boldsymbol{x}_i$ to particle $\boldsymbol{y}_j$ and $r$ denotes the Euclidean distance between $\boldsymbol{x}_i$ and particle $\boldsymbol{y}_j$. Each point is regarded as a particle, namely, each point has ideal conditions--mass but no volume or shape.}

%	$\mathbb{aaa},\mathbf{r}$,
	
	\emph{\textbf{Motion Representation:}} The motion for the template set $\mathbf{X}$ consists of a rotation and a translation, which is denoted as a tuple ($\textit{\textbf{p}}\in \revise{\mathbb{R}}^{3\times1},\textit{\textbf{t}}\in \revise{\mathbb{R}}^{3\times1}$).

	The rotation of the template set $\mathbf{X}$ is denoted by an axis-angle representation\revise{, which consist of a rotation vector $\textit{\textbf{p}}$ and a translation vector $\textit{\textbf{t}}$: 
	\begin{equation}
	\begin{aligned}\label{810}
	&\textit{\textbf{p}}=\theta_t\cdot\textit{\textbf{n}}_{p} \\
	&\textit{\textbf{t}}=s_t\cdot\textit{\textbf{v}}_{t}
	\end{aligned}
	\end{equation}}\revise{where $\theta_t$ represents the rotation angle, and $\textit{\textbf{n}}_{p}\in \revise{\mathbb{R}}^{3\times1}$ is the normalized rotation axis, $s_t$ represents the translation step, and $\textit{\textbf{v}}_{t}\in \revise{\mathbb{R}}^{3\times1}$ is the normalized direction of the translation.}
	
	\revise{Based on the per-point gravitation formulation, we present an illustration in Fig.~\ref{fig:motion}. The per-point gravitation $\boldsymbol{F}_{xi}$ is first decomposed into an axial translation force $\textit{\textbf{f}}_{i1}$ and a rotational force $\textit{\textbf{f}}_{i2}$:
		\begin{equation}
		\begin{aligned}
		&\boldsymbol{f}_{i 1}=\left(\boldsymbol{F}_{x i} \cdot \boldsymbol{n}_{x i}\right) \cdot \boldsymbol{n}_{x i} \\
		&\boldsymbol{f}_{i 2}=\boldsymbol{F}_{x i}-\boldsymbol{f}_{i \mathbf{1}}
		\end{aligned}
		\end{equation}
		where $\boldsymbol{n}_{x i}$ denotes the normalized point vector $\boldsymbol{x}_{i}$. The rotational force $\textit{\textbf{f}}_{i2}$ generates a torque $\textit{ \textbf{p}}_{xi}  $ for point cloud rotation that follows the right-hand rule of the law of force composition:
		\begin{equation}
		\textit{ \textbf{p}}_{xi}=\boldsymbol{x}_{i} \times \boldsymbol{f}_{i 2}
		\end{equation}
	Here, $\times$ represents the cross product of the vectors. The rotation axis $\textit{\textbf{n}}_{p}$ and the translation direction $\textit{\textbf{v}}_{t}$ are computed as follows:%%这里可以删
		\begin{equation}
		\begin{aligned}
		&\textit{\textbf{n}}_{p}=\frac{\sum_{i=1}^{\revise{N}} \textit{\textbf{p}}_{xi}}{\left\| \sum_{i=1}^{\revise{N}} \textit{\textbf{p}}_{x_i}\right\|} \\
		&\textit{\textbf{v}}_{t}=\frac{\sum_{i=1}^{\revise{N}} \textit{\textbf{f}}_{i1}}{\left\| \sum_{i=1}^{\revise{N}} \textit{\textbf{f}}_{i1}\right\|}
		\end{aligned}
		\end{equation}}

	where $\| \cdot \| $ denotes the vector length. More specifically, to apply the axis-angle representation $(\theta_t,\textit{\textbf{n}}_{p})$ to transform the template set, the rotation matrix $\textit{\textbf{R}}$ can be easily derived with the Rodrigues formula~\cite{Rf}: %这里也许能加个引用
		\begin{equation}\label{eqRf} 
		\textit{\textbf{R}}={\cos{\theta_t}}\cdot\textbf{I}+(1-\cos{\theta_t})\textit{\textbf{n}}_{p}\textit{\textbf{n}}_{p}^{T}+\sin\theta_t\cdot\tilde{\textit{\textbf{n}}}_{p}, \end{equation}
		where $\textbf{I}$ is a 3$\times$3 identity matrix and $\tilde{\textit{\textbf{n}}}_{p}$ denotes the skew-symmetric matrix of $\textit{\textbf{n}}_{p}$, which is formulated as:
		\begin{equation}\label{eqRfd} 
		\tilde{\textit{\textbf{n}}}_{p}=\left(\begin{array}{ccc}0 & -n_{p}^{z} & n_{p}^{y} \\ n_{p}^{z} & 0 & -n_{p}^{x} \\ -n_{p}^{y} & n_{p}^{x} & 0\end{array}\right), \end{equation}

	\begin{figure}[t!] 
		\center{\includegraphics[width=0.5\textwidth]{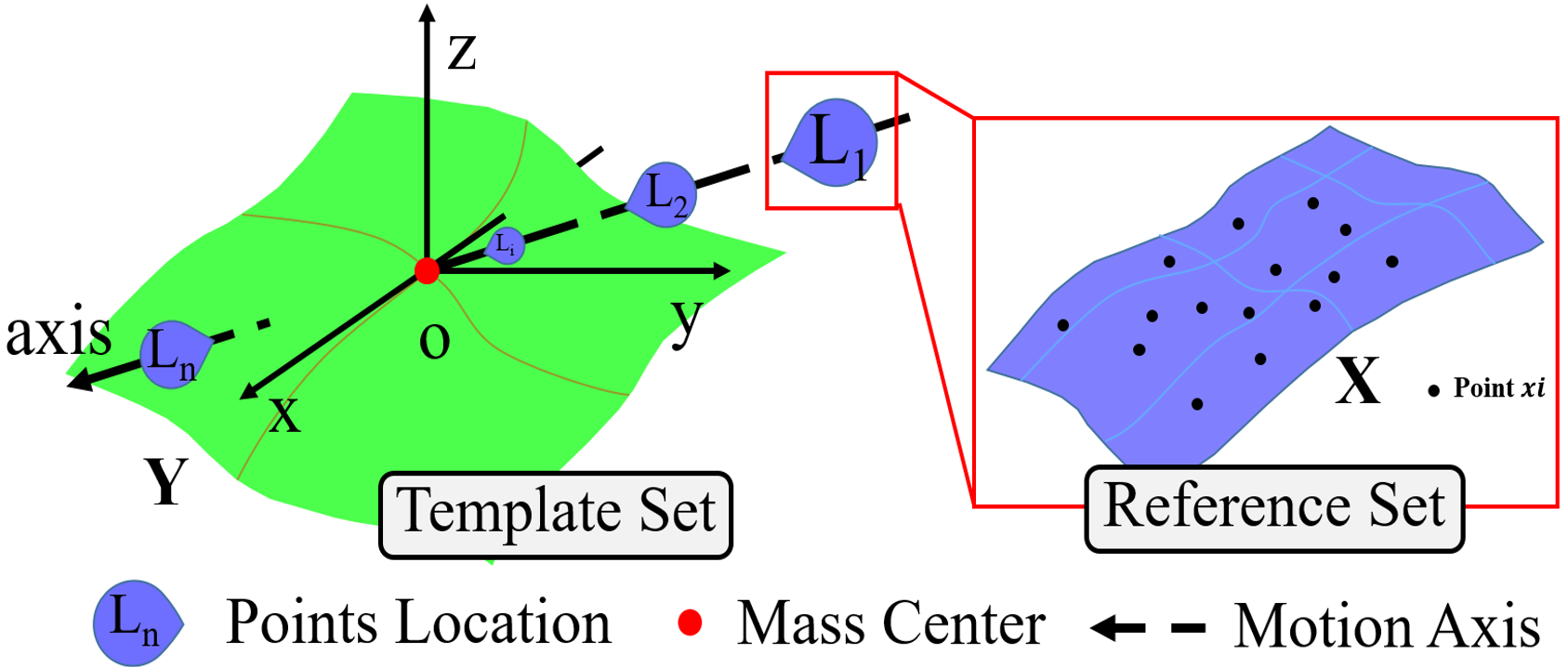}}
		\vspace{-5mm}
		\caption{\label{fig:crosspoint} Illustration of potential energy and gravitation cross functions with respect to linear motion. The green and blue surfaces represent the point sets of the reference and template, respectively. The blue water droplets $L$ move along the axis in 3D space and represent the different statuses of surface $\mathbf{X}$. The size of each water droplet indicates the level of potential energy, while the direction of the tail denotes the force traction.
		}
	\end{figure}
	
	\begin{figure}[t!] 
		\center{\includegraphics[width=0.5\textwidth]{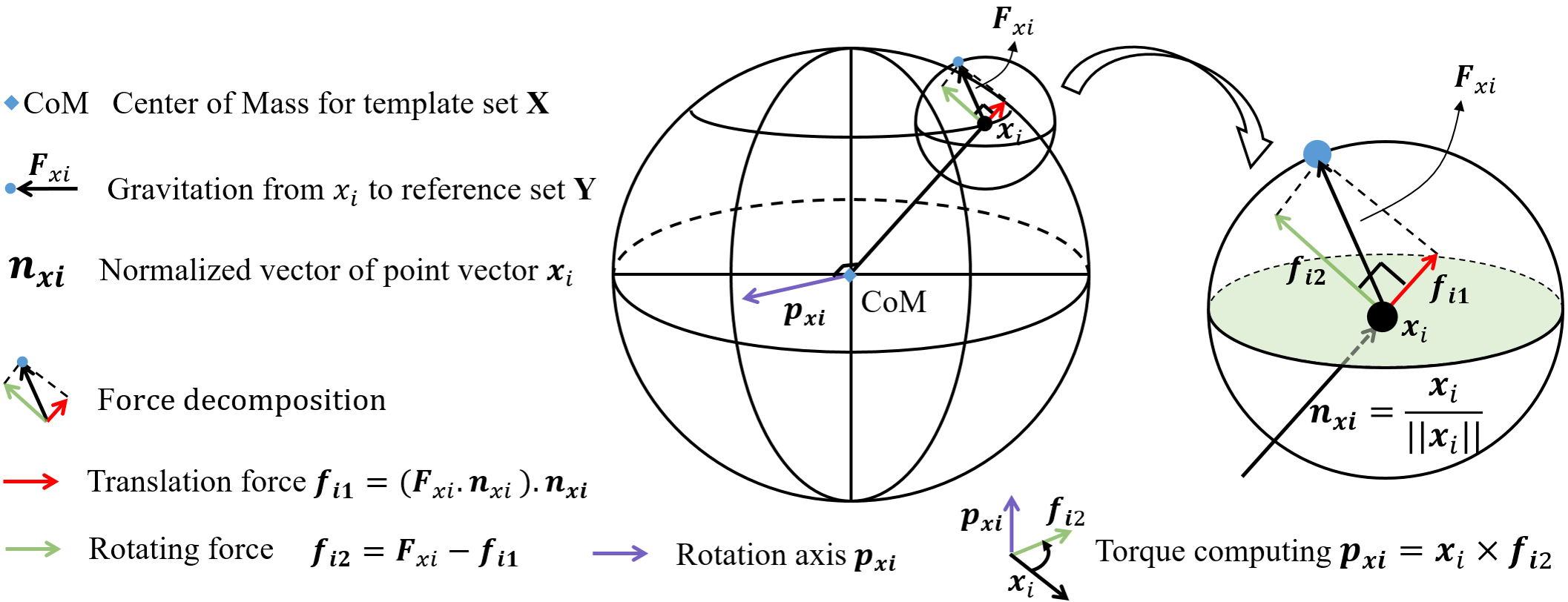}}
		\vspace{-5mm}
		\caption{\label{fig:motion} \revise{Illustration of the motion representation. The per-point gravitation is first decomposed into an axial translation force and a rotational force. The rotational force generates a torque for point cloud rotation.}
		}
	\end{figure}
	
	%%这里加个过度，前面的到了轴，现在要算步长。还要把花体的X改回来
	\revise{\emph{\textbf{Stride Control:}} With the rotation axis $\textit{\textbf{n}}_{p}$ and translation direction $\textit{\textbf{v}}_{t}$, we need to assign the appropriate stride, $\theta_t$ and $s_t$ to obtain the current transformation parameters.} For the stride computation, we apply a mutation detection scheme, which is illustrated for a one-dimensional movement for simplicity in Fig.~\ref{fig:crosspoint}. The template point set $\mathbf{X}$ (water droplets colored blue) moves along one axis in the gravitational field that is induced by the reference set $\mathbf{Y}$. When crossing the centroid o of $\mathbf{Y}$, the gravitation direction of $\mathbf{X}$ changes, and the potential energy of $\mathbf{X}$ is proportional to the distance between the centroid and $\mathbf{X}$. When the Euclidean distance $r$ approaches zero, the PE approaches a minimum, and at this moment, the point pair $(\boldsymbol{x}_i,\boldsymbol{y}_j)$ is also registered. However, the strides rarely approach the exact position of the optimal minimum, and we make some adjustments to the strides. Upon the mutation of PE, namely, when gravity is detected, we halve the strides.
	
	To capture the mutation, a pair of flags, namely, $F_R$ and $F_t$, are set to observe the changes in the rotational torque and gravitational vector, respectively. 
	In each iteration, we recompute $(\textit{\textbf{n}}_{p},\textit{\textbf{v}}_{t})$ and store its last values. The flags are defined as follows: \begin{equation}\label{RR}
	F_R=\textit{\textbf{n}}_{p}\cdot\textit{\textbf{n}}_{p(last)}, 
	\end{equation}
	\begin{equation}\label{tt}
	F_t=\textit{\textbf{v}}_{t}\cdot \textit{\textbf{v}}_{t(last)}.
	\end{equation}
	where $(\textit{\textbf{n}}_{p(last)},v_{t(last)} \revise{)}$ are the last iteration parameters of $(\textit{\textbf{n}}_{p},\textit{\textbf{v}}_{t})$.
	When the flag values are less than zero, the template set $\mathbf{X}$ has crossed the mutation point, and the force direction has reversed. Hence, the stride parameters ($\theta_t, s_t$) of rotation and translation are adjusted to approach the optimal values. Notably, the stride can be solved with other adjustment algorithms, whereas we only apply simple mutation-halving scheme, which is presented in detail in Algorithm~\ref{algorithm1}, to obtain a coarse alignment. When the stride parameters ($\theta_t, s_t$) are less than the thresholds ($\varepsilon_R,\varepsilon_t$), the iteration is stopped.
%%%TBD 0109
 
	The proposed MPE is a global registration method and can achieve accurate and efficient reconstruction. However, the limitation of MPE is that its time complexity exponentially rises with the number of points in the point clouds. For scalability to large point clouds, we employ a random downsampling strategy to obtain the same gravitational field distribution but lower intensity. The new sparse point clouds, which we denote as $\mathbf{X}_{d}\subset\mathbf{X}$ and $\mathbf{Y}_{d}\subset\mathbf{Y}$, increase the scalability and reduce the time complexity. However, the downsampling scheme also introduces sampling bias and is unable to achieve a perfect global minimum through the MPE method alone. Therefore, once the two point clouds are registered with the MPE method, we apply the trimmed iterative closest point algorithm to boost the convergence of registration. This coarse-to-fine approach is applied by many registration methods. However, these methods almost always rely on feature descriptors. In contrast, our proposed framework for global minimum approximation enables the point cloud registration procedure to cross the local minima of traditional $\ell_2$-based methods with a smoother optimization error map and directly approach the local convex optimization function under a natural force, without the need for feature descriptors.

	\begin{figure}[t!] 
	\center{\includegraphics[width=0.5\textwidth]{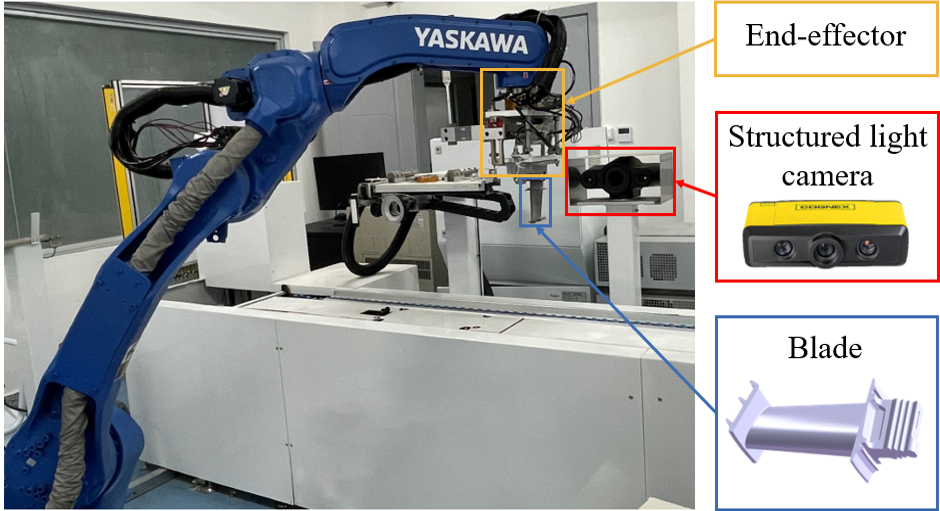}}
	\vspace{-5mm}
	\caption{\label{fig:fig4} Reconstruction system.}
	\vspace{-5mm}
\end{figure}
\begin{figure*}[t!] 	
	\centering
	\begin{tikzpicture}[inner sep=0pt,outer sep=0pt]
	\node[anchor=south west] (A) at (0in,0in)
	{\includegraphics[width=.99\textwidth,clip=false]{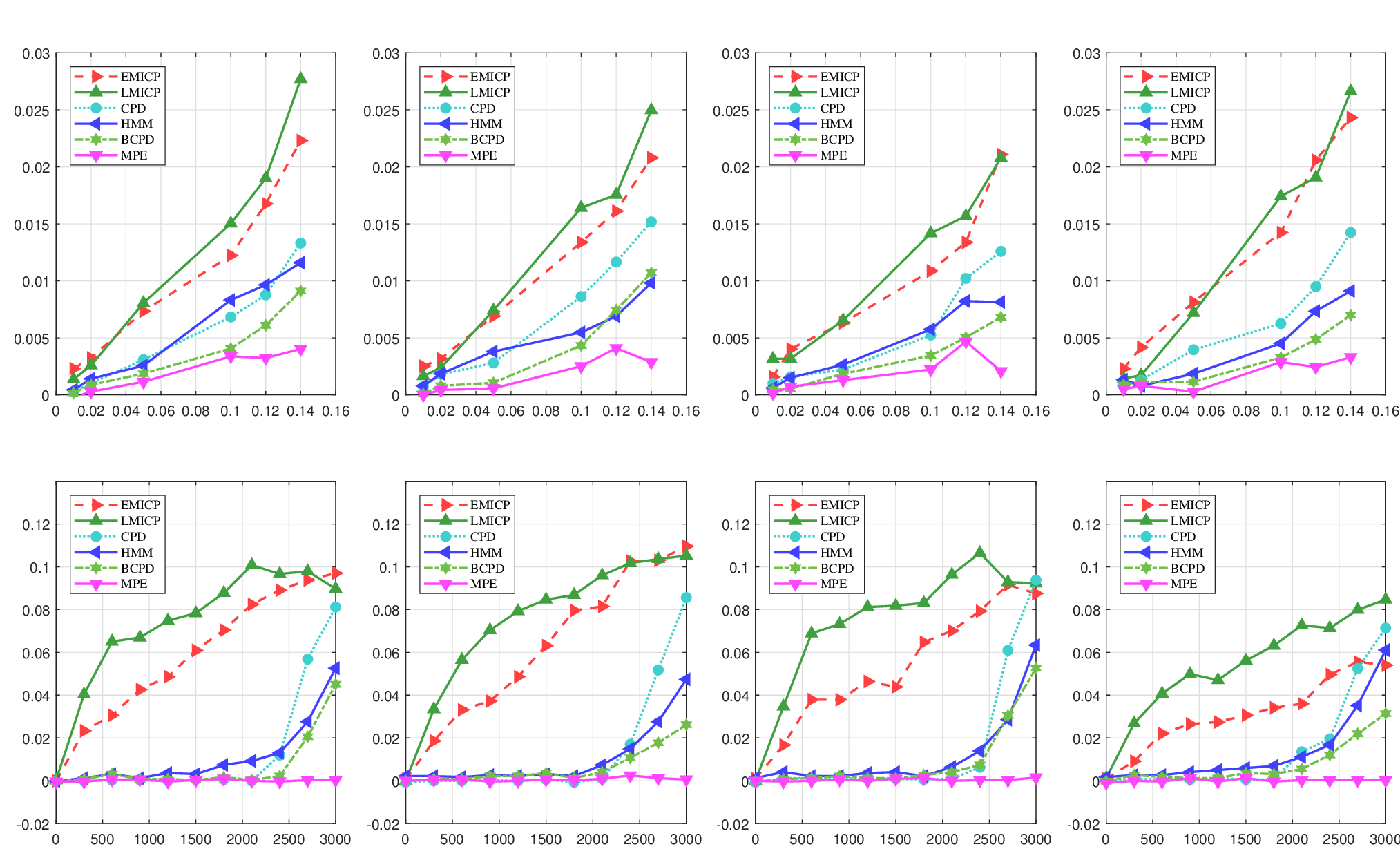}};
	
	\node[anchor=north,xshift=-183pt,yshift=155pt] at (A.south)
	{\footnotesize (a) Gaussian Noise (m)};
	\node[anchor=north,xshift=-54pt,yshift=155pt] at (A.south)
	{\footnotesize (b) Gaussian Noise (m)};
	\node[anchor=north,xshift=74pt,yshift=155pt] at (A.south)
	{\footnotesize (c) Gaussian Noise (m)};
	\node[anchor=north,xshift=201pt,yshift=155pt] at (A.south)
	{\footnotesize (d) Gaussian Noise (m)};
	
	\node[anchor=north,xshift=-1pt,yshift=-90+8pt,rotate=90] at (A.west)
	{\footnotesize RMSE (m)};
	\node[anchor=north,xshift=-1+127.5pt,yshift=-90+8pt,rotate=90] at (A.west)
	{\footnotesize RMSE (m)};
	\node[anchor=north,xshift=-1+127.5*2pt,yshift=-90+8pt,rotate=90] at (A.west)
	{\footnotesize RMSE (m)};
	\node[anchor=north,xshift=-1+127.5*3pt,yshift=-90+8pt,rotate=90] at (A.west)
	{\footnotesize RMSE (m)};
	\node[anchor=north,xshift=-183pt,yshift=-1pt] at (A.south)
	{\footnotesize (e) Number of Outliers (\#pts)};
	\node[anchor=north,xshift=-54pt,yshift=-1pt] at (A.south)
	{\footnotesize (f) Number of Outliers (\#pts)};
	\node[anchor=north,xshift=74pt,yshift=-1pt] at (A.south)
	{\footnotesize (g) Number of Outliers (\#pts)};
	\node[anchor=north,xshift=201pt,yshift=-1pt] at (A.south)
	{\footnotesize (h) Number of Outliers (\#pts)};
	
	\node[anchor=north,xshift=-1pt,yshift=90-8pt,rotate=90] at (A.west)
	{\footnotesize RMSE (m)};
	\node[anchor=north,xshift=-1+127.5pt,yshift=90-8pt,rotate=90] at (A.west)
	{\footnotesize RMSE (m)};
	\node[anchor=north,xshift=-1+127.5*2pt,yshift=90-8pt,rotate=90] at (A.west)
	{\footnotesize RMSE (m)};
	\node[anchor=north,xshift=-1+127.5*3pt,yshift=90-8pt,rotate=90] at (A.west)
	{\footnotesize RMSE (m)};
	
	\end{tikzpicture}

	\vspace{-3mm}
	\caption{\label{fig:fig2} Performance comparison with several state-of-the-art algorithms using RMSE metrics. The first row: Registration errors of four blades with Gaussian noise with standard deviations of 0.01 to 0.14 (m). The second row: Registration errors with different numbers of outliers that are generated from a uniform model. The columns present the results of models (a), (b), (c), and (d) from left to right. }
\end{figure*}
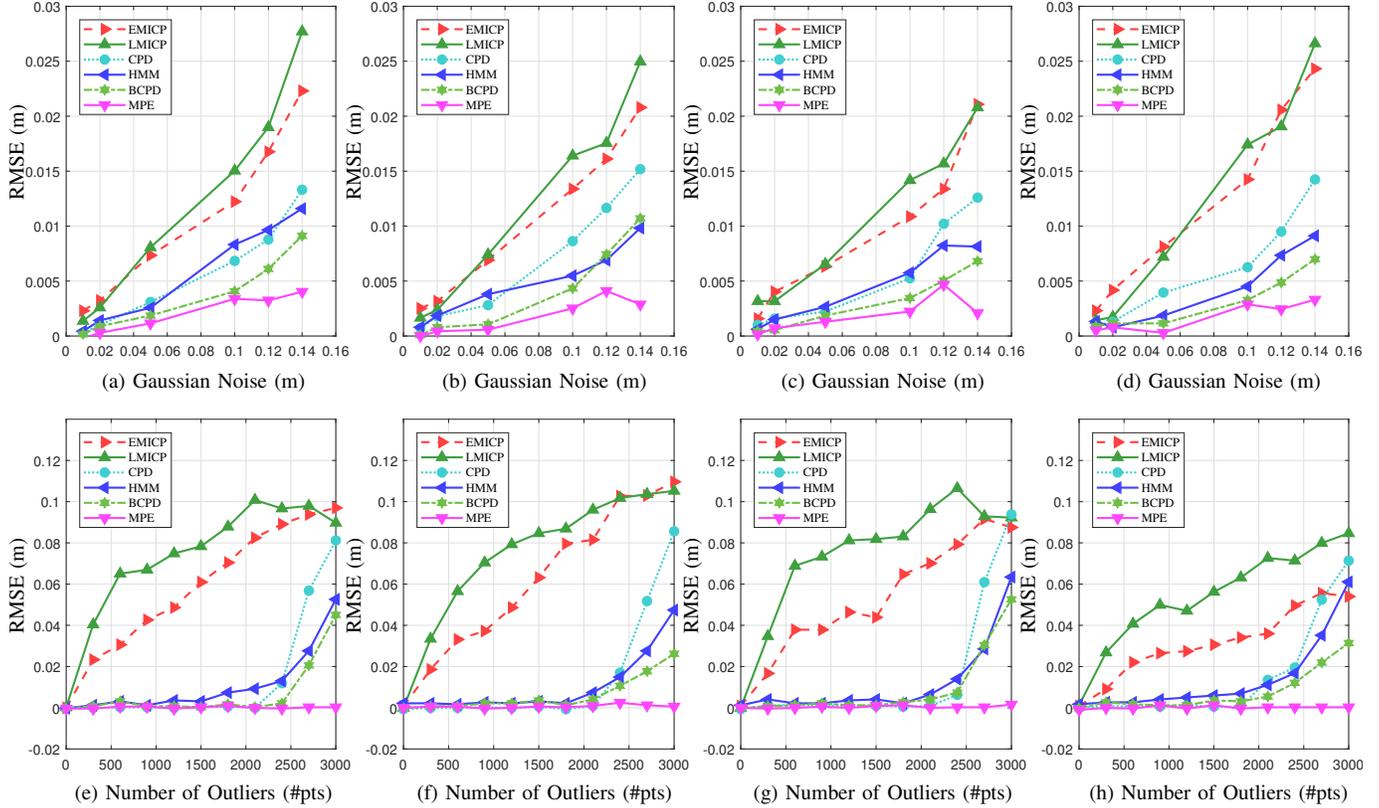
	
	\section{Experiments}
\begin{figure}[t!] 
	\center{\includegraphics[width=0.5\textwidth]{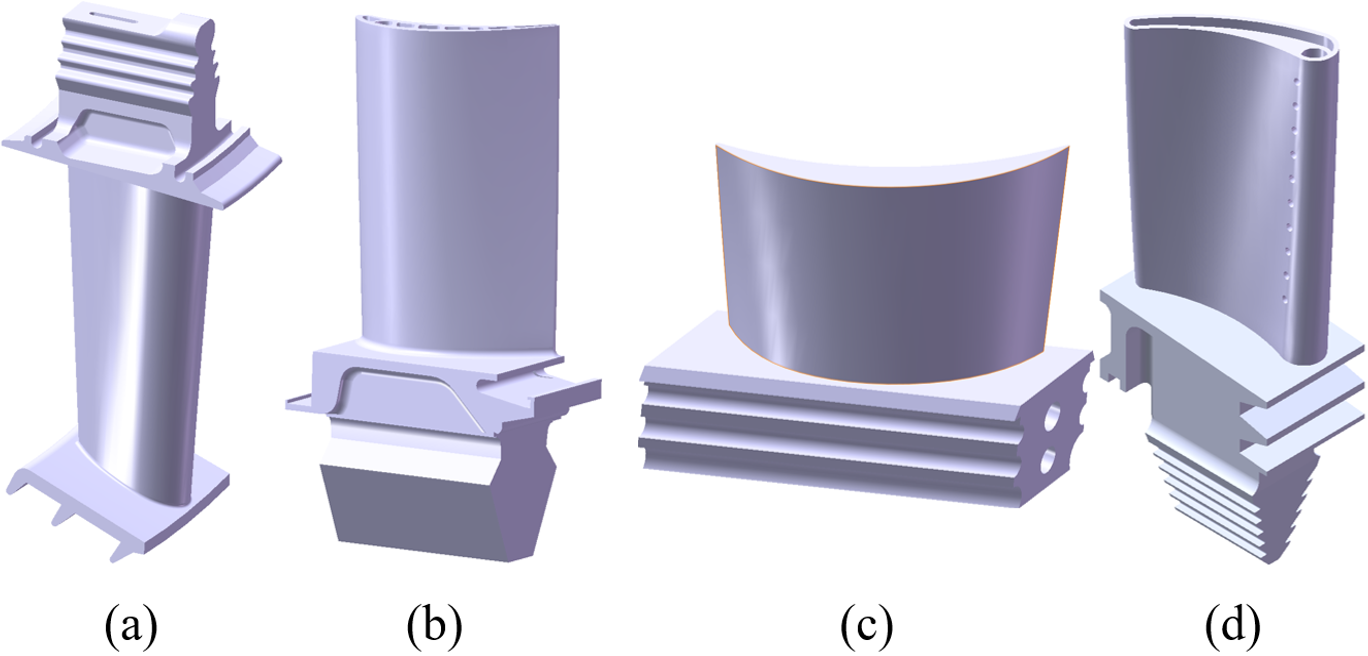}}
	\vspace{-5mm}
	\caption{\label{fig:fig1} CAD models (1-4) of four turbine blades: Model 1 (a) is a low-pressure turbine blade, while models (2-4)(b)(c)(d) are different types of aero-engine high-pressure turbine blades.}
\end{figure}

	The proposed method is implemented on a PC with an Intel i5 3.4 Hz processor, and a series of experiments are designed to test its performance on different data sets.
	We verify the theoretical properties of our method using idealized 3D blade models, which are shown in Fig.~\ref{fig:fig1}. The proposed method is compared with other algorithms, and the influence of the hyperparameter settings on our method is discussed. Furthermore, we introduce our hardware reconstruction system and visualize the registration procedure for four real blades. Specifically, we compare the  proposed MPE method and other state-of-the-art methods in terms of accuracy. 
% blade model picture (a) ?????? (b) ?????? (c) 

	\subsection{Performance Evaluation}
In this section, we demonstrate the robustness and convergence properties of the proposed method and compare the results with those of other algorithms.

\begin{figure}[t!] 
	\center{\includegraphics[width=0.5\textwidth]{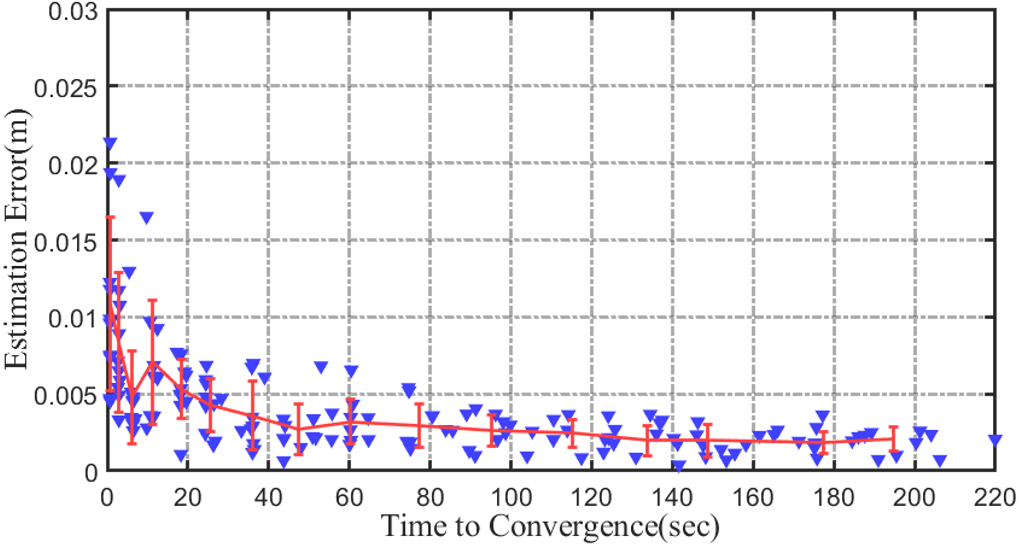}}
	\vspace{-5mm}
	\caption{\label{fig:ET} Estimation error and convergence time results with different sampling ratios. Every blue inverted triangle denotes a data point that represents the relation between the estimation error and convergence time under the corresponding sampling ratio. The red line represents their averages and the variance of the estimation error with increasing sampling ratio. The metrics are calculated from 160 runs with 16 sampling ratios.}
\end{figure}

\begin{figure*}[t!]

	\centering
	\begin{tikzpicture}[inner sep=0pt,outer sep=0pt]
	\node[anchor=south west] (A) at (0in,0in)
	{\includegraphics[width=.99\textwidth,clip=false]{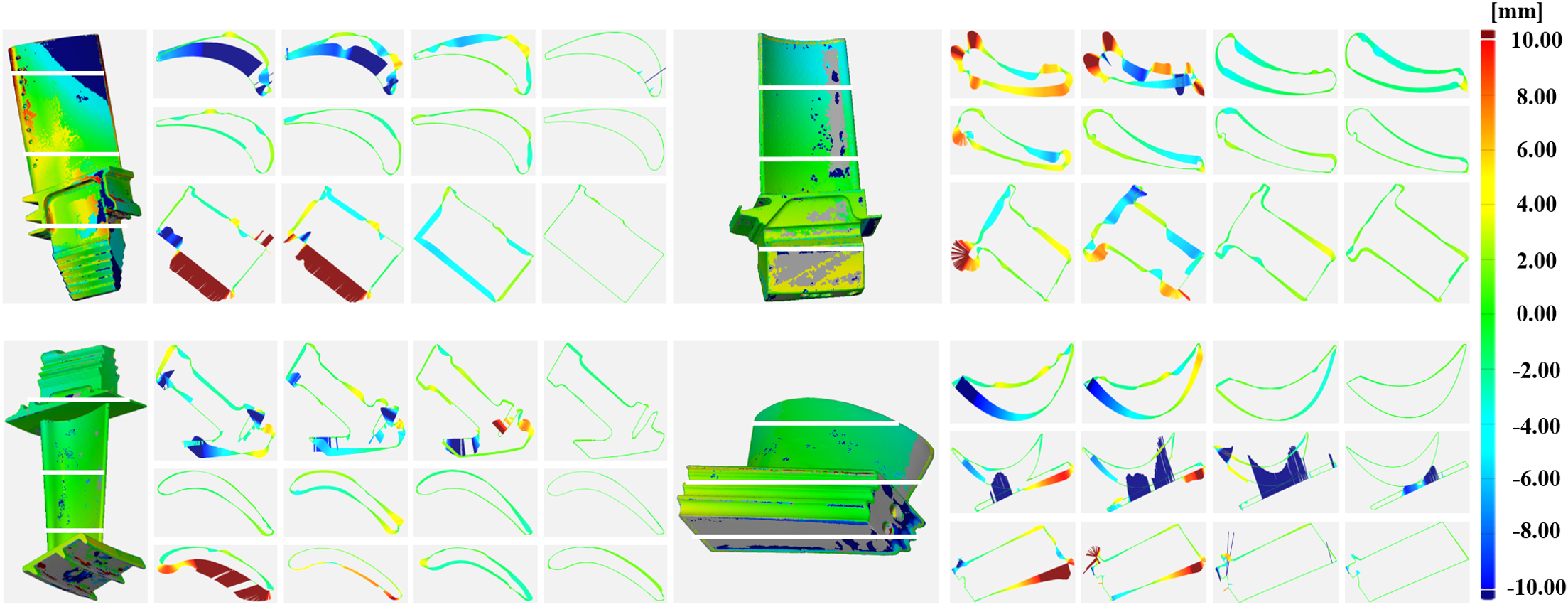}};
	
	\node[anchor=north,xshift=-186pt,yshift=92pt] at (A.south)
	{\footnotesize (a)};
	\node[anchor=north,xshift=75pt,yshift=92pt] at (A.south)
	{\footnotesize (b)};
	\node[anchor=north,xshift=-186pt,yshift=-2pt] at (A.south)
	{\footnotesize (c)};
	\node[anchor=north,xshift=75pt,yshift=-2pt] at (A.south)
	{\footnotesize (d)};
	
	\end{tikzpicture}

	\vspace{-4mm}
	\caption{\label{fig:fig511} Cross-sectional error chromatogram of blade reconstruction. The columns correspond to three cross-sections for four methods, namely, ICP, CPD, BCPD, and MPE, from left to right in order. The white lines on the blades represent the locations of the three cross-sections. }
\end{figure*}

\emph{$\textit{1)}$ Robustness of the Proposed Algorithm: } Using the four blade models, we compare our method to the following baselines: the coherent point drift (CPD)~\cite{CPD}, LM-ICP (Levenberg-Marquart ICP)~\cite{OT1}, EM-ICP (expectation-maximization ICP)~\cite{EMICP}, hybrid mixture model (\cite{HMM}), and Bayesian formulation of coherent point drift (~\cite{BCPD}) algorithms. The size of each model is adjusted to 3$\times$3000 through a down-sampling scheme to increase the effectiveness of the test. We use a random translation and rotation (angle $\phi\in [0,\frac{\pi}{2}]$). Then, we introduce Gaussian noise that is generated by a Gaussian distribution into the template set Y.

Fig.~\ref{fig:fig2}(a-d) shows that the proposed MPE algorithm outperforms the five baseline methods. With increasing levels of noise, the registration error of our method remains the lowest, and the slope of the curve also remains relatively small. The reference point set is generated with different levels of noise.
The translation of every point from the original position is randomly computed by a Gaussian model in 3D space. The standard deviations of the Gaussian distribution are $\sigma=0.01-0.14$ (m). As shown in the first row of Fig.~\ref{fig:fig2}, our algorithm achieves the lowest error among all algorithms by at least $45.50\%$. More interestingly, our method shows less error accumulation for all Gaussian noise levels.

Fig.~\ref{fig:fig2}(e-h) compares our method with the baselines in terms of RMSE with an increasing number of outliers (uniform noise). The outliers ($0 \to 3000$ points) are generated uniformly in a cube that encompasses the point cloud data. The proposed MPE algorithm performs exceptionally well and outperforms CPD and LMICP. Among the three GMM-based methods, BCPD outperforms HMM and CPD. However, it does not show large differences in blade model tests. Our method shows robustness to a large number of outliers by taking advantage of the global PE minimum to ignore the gravitational field that is induced by the outliers. The estimation error always stays at a low value that approaches zero in this experiment irrespective of the number of outliers.
Fig.~\ref{fig:fig2}(e-h) demonstrates the great registration effectiveness with the increasing number of outliers. The reference point set is generated by adding random outliers in the data cube following a uniform distribution. The numbers of noise points are $200$, $950$, $2000$ and $2800$.

	\emph{$\textit{2)}$ Time to Convergence:} The next experiment is designed to examine the influence of the hyperparameter settings of the search on the tradeoff between convergence speed and reconstruction accuracy. To register point clouds with our proposed method, a downsampling rate generally must be specified. A higher sampling rate usually results in not only higher accuracy but also longer runtime. Therefore, to maintain the balance between the downsampling size of MPE computation and algorithm accuracy, we design an independent experiment to obtain 160 data points of multiple test results from experiments that are repeated 10 times. 
	In each experiment, we set the downsampling rate from 5\% to 80\% with an interval of 5\%. Then, we measure the time to convergence and calculate the error for each downsampling ratio. This process is visualized in Fig.~\ref{fig:ET}. The blue inverted triangle points are the original test data points, whereas the red line represents the time average for each downsampling ratio and its corresponding standard deviation. Table~\ref{tab:MPEL} presents the time cost of the MPE method under the different numbers of sampling points.  \revise{The times to convergence with different sampling ratios are presented in detail in Table~\ref{tab:tc}.}
	Obviously, as the sampling ratio increases, the estimation error decreases and becomes more stable. Significantly, the error line becomes smoother when the sampling ratio value is over 0.25, and our algorithm can ensure that a global optimum is attained more frequently.
	
	\begin{table*}[htbp]
	\centering
	\addtolength{\tabcolsep}{-2pt}
	\caption{\label{tab:tc} \revise{Average error and error variance of time to convergence with different downsampling ratios}}
	\begin{tabular}{c| c |c| c |c |c |c| c| c | c| c| c| c| c| c |c |c }
		\toprule
		Downsampling ratio & 0.05 & 0.10 & 0.15&0.20&0.25&0.30&0.35&0.40&0.45&0.50&0.55&0.60&0.65&0.70&0.75&0.80\\
		\midrule
		Average error (mm)	& 10.8 & 8.3 & 4.8&7.1&\textbf{5.3}&4.3&3.5&2.7&3.2&2.9&2.6&2.4&1.9&1.9&1.8&1.8	\\
		Error variance ($\times 10^{-6}$) &  32.1& 20.7&8.9&16.3&\textbf{3.67}&3.03&4.85&2.75&2.18&2.07&1.07&0.78&0.94&0.97&1.13&0.49\\
		\bottomrule
	\end{tabular}
\end{table*}

\begin{table}[htbp]
	\centering
	\caption{\label{tab:MPEL} \revise{Time cost of minimum PE registration for naive MPE theory with the full sample and two samples that have been downsampled at different rates}}
	\begin{tabular}{c| c c c}
		\toprule
		N$\times$M & \revise{full sample} & \revise{200 pts} & \revise{100 pts}\\
		\midrule
		500$\times$3	& 16.4s & 3.3s & 0.9s\\
		2000$\times$3	& 5.1m & 3.9s & 1.5s\\ 
		8000$\times$3	& 1.6hr & 7.2s & 4.8s\\ 
		32000$\times$3	& 30.6hr & 20.2s & 18.4s\\ 
		\bottomrule
	\end{tabular}
\end{table}
	
	%%%%%%%%%%%%%%%%%%%%%%%%%%%%%%%%%%%%%%%%%%%%%%%%%%%%%%%%%%%%%%%%%%%%%%%%这个表加描述
	
	\revise{\emph{$\textit{3)}$ Ablation Study on the Proposed Components:} We show the contributions of various module selections of our proposed MPE method, which consists of the down-Sample scheme (DS), motion control (MC), and ICP refinement. The performances of various combinations are reported in Table~\ref{tab:ab}. The models are still normalized and adjusted to 3000 pts. The standard deviation of the Gaussian distribution is set to $\sigma=20$ (mm), and we add $10\%$ outliers that are uniformly distributed in the domain $[-1,1]^{3}$.}
	
	\revise{Each combination is repeated ten times, and their average is taken as the final result. The performances with accuracy and runtime are evaluated in this experiment. The first part in the top three rows of Table~\ref{tab:ab} consists of three combinations with different downsampling rates, all points/full samples, 500 points samples, and 100 points samples. The DS+MC combination can represent the main part of the MPE algorithm. It is obvious that the accuracy increases as the downsampling rate increases; However, the runtime also increases. Hence, ICP refinement is considered to boost the convergence in the rows (4-6) of Table~\ref{tab:ab}. With the ICP module, the MPE (DS+MC+ICP) algorithm approaches the same level of accuracy under the 500 point samples, but with only 11.33\% of the DS+MC runtime. For a fair comparison, we also present the performance of pure ICP method. Because the DS+ICP combination is nonsense, the DS scheme only brings worse performance to ICP method. In conclusion, the DS module reduces the average runtime but results in lower accuracy. The ICP module is applied to solve the problem of decreasing accuracy. More importantly, only with the motion control algorithm is the registration accurate and efficient.}
	
	\revise{As shown in Table~\ref{tab:ab}, the MPE (DS+MC+ICP) algorithm shows great performance and efficiency for registration. This combination will be regarded as the standard paradigm for our MPE method and, hence, applied to multiview registration for blade measurement in practice.}
	
	\begin{table}[htbp]
		\centering
		\caption{\label{tab:ab} \revise{Impact of module selection}}
		\begin{tabular}{c c c | c c}
			\toprule
			DS 		& MC		&	ICP	&	Accuracy/mm	& Runtime/s\\
			\midrule
			Full	&	\gou	&\wu 	&	0.12		&	241.54	\\
			500 pts	&	\gou	&\wu 	&	5.14		&	15.34	\\
			100 pts	&	\gou	&\wu 	&	10.03		&	1.94\\
			\midrule
			Full	&	\gou	&\gou	&	0.10		&	276.41	\\
			500 pts	&	\gou	&\gou	&	\textbf{0.11}		&	\textbf{18.64}	\\
			100 pts	&	\gou	&\gou	&	10.04		&	2.32\\
			\midrule
			Full	&	$-	$	&\gou	&	60.43		&	22.54\\
			500 pts	&	$-	$	&\gou	&	125.32		&	3.18\\
			100 pts	&	$-	$	&\gou	&	230.14		&	0.93\\
			\bottomrule
		\end{tabular}
	\end{table}
%       $\checkmark$ & $\checkmark$ & $-	$\\
	
	\begin{table}[htbp]
		\centering
		\caption{\label{tab:UWAe2} Average cross-sectional errors ($mm$)  of four models}
		\begin{tabular}{c | c c c c}
			\toprule
			Blade  & Model\_1	&	Model\_2	&	Model\_3	& Model\_4\\
			\midrule
			EMICP	&	0.48	&0.65&	0.50&	0.69	\\
			CPD	&	0.24	&0.38&	0.33&	0.59	\\
			BCPD	&	0.13	&0.16&	0.17&	0.21\\
			MPE		&	\textbf{0.04}	&\textbf{0.12}	&	\textbf{0.08}	&	\textbf{0.09}\\
			\bottomrule
		\end{tabular}
	\end{table}

		\begin{table}[htbp]
		\centering
		\addtolength{\tabcolsep}{-2pt}
		\caption{\label{tab:time} \revise{Average per-scan computation times ($sec$) of four models for registration}}
		\begin{tabular}{c | c c c c | c}
			\toprule
			blade  & Model1	&	Model2	&	Model3	& Model4 & Average\\
			\midrule
			EMICP	&	1013.27	&986.45&	1146.32&	1025.65	& 1042.92\\
			CPD		&	5698.32	&6215.49&	6389.80&	4432.15	& 5683.94\\
			BCPD	&	423.58	&544.35&	521.33&	468.78		& 489.51\\
			MPE		&	\textbf{391.84}	&\textbf{449.48}	&	\textbf{406.72}	&	\textbf{387.40}&\textbf{408.86}\\
			\bottomrule
		\end{tabular}
	\end{table}

	%\revise{\emph{$\textit{3)}$ Comparison with the maker point reconstruction:} }
	
	\subsection{Multiview registration with the blade measurement system} The blade measurement system is introduced in Fig.~\ref{fig:fig4}. The whole system consists of an end effector that is equipped at the end of a Yaskawa DX200 6-axis robot and a Cognex ES-A5000 binocular structured light camera that is fixed on the platform. The scanned blades are grabbed by the end effector, which generates different blade poses to be scanned with the fixed camera. In this experiment, the poses of the blade are generated by z-axis rotation with an interval of 10 degrees and random (x,y)-axis deviation at 45\degree. 

	We demonstrate the performance of our proposed registration algorithm in multiview reconstruction and realize the registration of different views by processing the raw point clouds of four objects. Due to the absence of ground-truth data, we use the rotation error $\epsilon_r$ and translation error $\epsilon_t$ to measure the performance of the reconstruction result. The rotation error $\epsilon^{ij}_r$ and translation error $\epsilon^{ij}_t$ are calculated as follows:
\begin{equation}\label{eq11}  \epsilon^{ij}_r=\frac{180}{\pi}*arcos(\frac{trace(\textit{\textbf{R}}_{GT}^{ij}*(\textit{\textbf{R}}_{E}^{ij})^{-1})-1}{2})\end{equation}
\begin{equation}\label{eq12}  \epsilon^{ij}_t=\|\textit{\textbf{t}}_{GT}^{ij}-\textit{\textbf{t}}_{E}^{ij} \| \end{equation}
$\textit{\textbf{R}}_{GT}^{ij}$ and $\textit{\textbf{t}}_{GT}^{ij}$ represent the ground-truth rotation and translation, respectively, between the i-th and j-th point sets. The point clouds are coarsely aligned manually and then refined with the ICP algorithm for ground-truth generation. $\textit{\textbf{R}}_{E}^{ij}$ and $\textit{\textbf{t}}_{E}^{ij}$ denote the estimated rotation and translation, respectively. 
	
%	\begin{table*}[htbp]
%		\centering
%		\caption{\label{tab:UWAe1}Registration Error (Rotation(\degree)/Translation($mm$))  of Four Models}
%		\begin{tabular}{c c c c c c}
%			\toprule
%			\diagbox{Method}{Error}{Model}&Model1	&	Model2	&	Model3	& Model4\\
%			\midrule
%			Trimmed-icp~\cite{Trimmedicp}	&	5.3751/0.69	&	5.0682/0.42	&	1.9188/0.71	&	1.6655/0.39\\
%			RoPs~\cite{algo2}	&	2.1127/0.21	&	2.5623/0.34	&	1.1633/0.16	&	2.9276/0.22	\\
%			RoPs+generalized-ICP ~\cite{algo1}~\cite{gedicp}	&	0.9311/0.14	&	1.2313/0.16	&	0.7871/0.12	&	1.4516/0.23	\\
%			MPE		&	\textbf{0.2651}/\textbf{0.06}	&	\textbf{0.3865}/\textbf{0.05}	&	\textbf{0.1527}/\textbf{0.03}	&	\textbf{0.3230}/\textbf{0.04}\\
%			\bottomrule
%		\end{tabular}
%	\end{table*}
	
	\begin{table*}[htbp]
		\centering
		\addtolength{\tabcolsep}{8pt}
		\caption{\label{tab:UWAe1}Registration Error (Rotation (\degree)/Translation($mm$)) of Four Models}
		\begin{tabular}{c |c c c c c}
			\toprule
			\diagbox{Method}{Error}{Model}&Model1	&	Model2	&	Model3	& Model4\\
			\midrule
			EMICP~\cite{EMICP}	&	4.81/0.60	&	4.71/0.41	&	3.17/0.67	&	2.37/0.42\\
			CPD~\cite{CPD}	&	1.77/0.31	&	2.21/0.22	&	1.43/0.43	&	1.98/0.29	\\
			BCPD ~\cite{BCPD}	&	0.87/0.13	&	1.24/0.12	&	0.65/0.16	&	1.23/0.26	\\
			MPE		&	\textbf{0.27}/\textbf{0.06}	&	\textbf{0.38}/\textbf{0.05}	&	\textbf{0.15}/\textbf{0.03}	&	\textbf{0.32}/\textbf{0.04}\\
			\bottomrule
		\end{tabular}
	\end{table*}
	
	\begin{figure}[t!] 	
		\centering
		\begin{tikzpicture}[inner sep=0pt,outer sep=0pt]
		\node[anchor=south west] (A) at (0in,0in)
		{\includegraphics[width=.5\textwidth,clip=false]{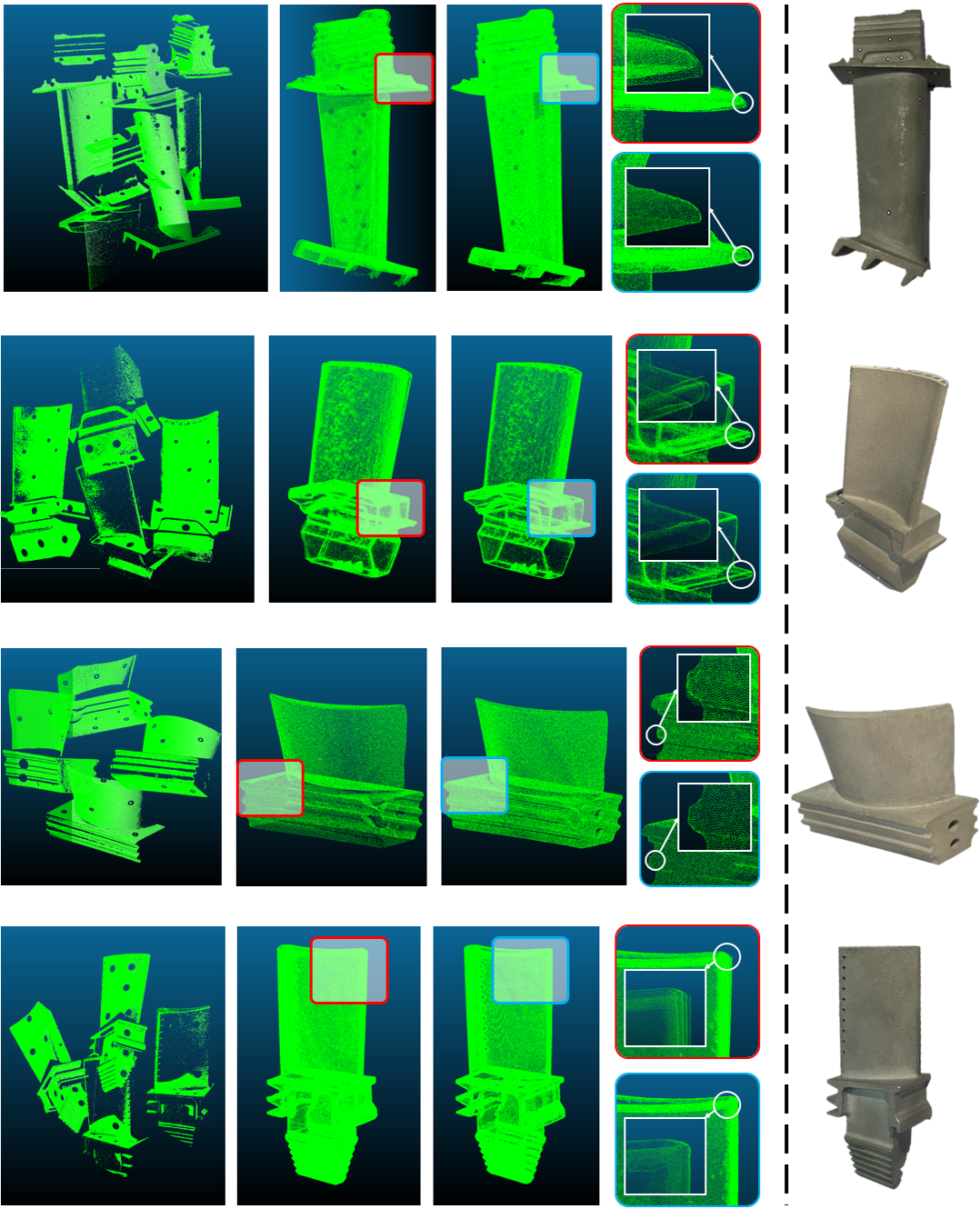}};
		\node[anchor=north,xshift=-10pt,yshift=238pt] at (A.south)
		{\footnotesize (a)};
		\node[anchor=north,xshift=-10pt,yshift=157pt] at (A.south)
		{\footnotesize (c)};
		\node[anchor=north,xshift=-10pt,yshift=83pt] at (A.south)
		{\footnotesize (e)};
		\node[anchor=north,xshift=-10pt,yshift=-1pt] at (A.south)
		{\footnotesize (g)};
		
		\node[anchor=north,xshift=104pt,yshift=238pt] at (A.south)
		{\footnotesize (b)};
		\node[anchor=north,xshift=104pt,yshift=157pt] at (A.south)
		{\footnotesize (d)};
		\node[anchor=north,xshift=104pt,yshift=83pt] at (A.south)
		{\footnotesize (f)};
		\node[anchor=north,xshift=104pt,yshift=-1pt] at (A.south)
		{\footnotesize (h)};
		\end{tikzpicture}

		\vspace{-3mm}
		\caption{\label{fig:fig52} Illustration of multiview reconstruction for blades. (a)(c)(e)(g) present the reconstruction demonstrations of models 1-4. Each part consists of four figures, which show the initialization, coarse alignment, fine alignment, and local details of coarse and fine alignment. The top red and bottom blue squares in the local details represent locally enlarged images that are sampled from the coarse and fine alignments, respectively. (b)(d)(f)(h) represent real models 1-4.} 
	\end{figure}
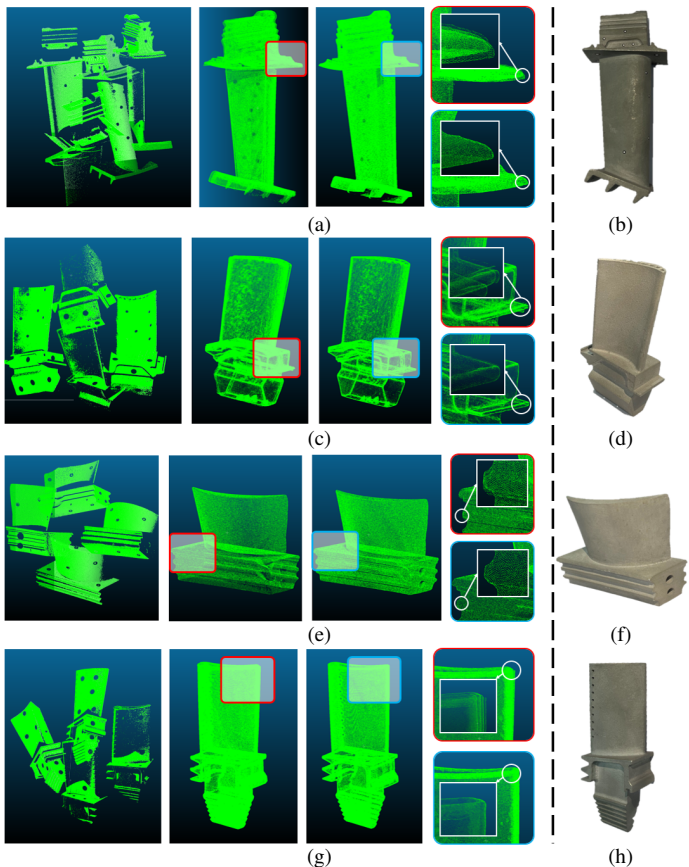 %%不用fig
	
	We compare our method with other methods, namely, ICP~\cite{icp1}, coherent point drift~\cite{CPD}, and Bayesian coherent point drift~\cite{BCPD}, in terms of rotation and translation errors in Table~\ref{tab:UWAe1}. For the multiview registration and input of a set of point clouds {$P_{1},P_{2},......,P_{N},$} sequentially, the algorithm starts with two views {$P_{1},P_{i+1}$} (i=1 initially) and performs the registration. For a registered new point cloud $Q_{new}$ that is generated by our proposed method, we regard {$P_{1}, P_{i+1}$} as having been registered successfully if the number of overlapping points exceeds a threshold ${\tau}_{0}$. We update the registered new point cloud $Q_{new}$ as the new $P_{1}$. Otherwise, the latter input point cloud is discarded, and the registration is continued iteratively until all point sets are processed. Furthermore, to reduce computation, we merge the registered pairwise points into a single point after accepting the successful registration result. As presented in Table~\ref{tab:UWAe1}, our proposed method achieves the best registration performance and shows great potential in free-form blade reconstruction. 
	For a better comparison, we evaluate the registration performance based on the multiview reconstruction cross-section, and the error chromatogram is shown in Fig.~\ref{fig:fig511}. We consider three cross-sections of each blade and mark their locations with white lines. The color and length of each vector denote the error level. Moreover, the columns present the results of ICP, CPD, BCPD, and MPE from left to right. Additional details are presented in Table~\ref{tab:UWAe2}. \revise{In addition to the accuracy in the real blade performance comparison, the average per-scan computation time for registration is presented in Table~\ref{tab:time}. The CPD algorithm shows the highest runtime. The large scale of the blade point cloud exponentially increases the computation time. The CPD algorithm reaches the maximum number of iterations and stops, which might be the main reason for the highest time cost. BCPD is more efficient but not as fast enough as MPE. Our MPE method can deal with the large number of points in each scan with a low computation time.}
	
	An overall visual illustration of multiview registration of four blades by our algorithm is presented in Fig.~\ref{fig:fig52}. Each row presents a procedure of single blade reconstruction with initialization, coarse alignment, fine alignment, local details, and the real model image from left to right. The top red and bottom blue squares present the coarse and fine alignment details, respectively. For better visualization, we only show four view scans in initialization instead of all scans.
	Interestingly, the MPE method shows a satisfactory overall result, whereas the enlarged image shows that it still has not approached the optimal result with a finite number of iterations without ICP refinement. Due to the exponential growth of gravitation as the distance decreases, there is an oscillation around the optimum at the end of the reconstruction procedure. Hence, it is more efficient and accurate to use ICP refinement than to increase the number of MPE iterations.
	In this experiment, we compare the performance of our proposed method with those of several other methods on real blade reconstruction. The results show that our algorithm achieves the best performance in blade reconstruction.

	\section{Conclusion}
	In this paper, we proposed a novel, globally optimal theory for free-form blade reconstruction in 3D, which was achieved by introducing a suitable physical system to approach the MPE of the point clouds. We also proposed a new NFI criterion to address the overfitting problem. Our method of gathering the point clouds using the rotational torque and gravitational vector efficiently realized globally optimal registration. The proposed approach was able to align point clouds within large amounts of Gaussian and uniform noise while running with low time complexity. According to a comparison between our proposed method and other approaches, our approach performed well. Different from the traditional baseline methods, our proposed method explores a new path to offer a new solution.

	\appendices

	\ifCLASSOPTIONcaptionsoff
	\newpage
	\fi
	
	\bibliographystyle{IEEEtran}
	\bibliography{Reference}

\end{document}